\pdfoutput=1

\documentclass[11pt]{article}

\usepackage{EMNLP2023}

\usepackage{times}
\usepackage{latexsym}

\usepackage[T1]{fontenc}

\usepackage[utf8]{inputenc}

\usepackage{microtype}
\usepackage{hyperref}
\usepackage{inconsolata}

\usepackage{amsmath} 
\usepackage{graphicx}
\usepackage{booktabs}
\usepackage{multirow} 
\usepackage{arydshln}
\usepackage{makecell}
\usepackage{colortbl}
\definecolor{mygray}{gray}{.9}

\title{PSST: A Benchmark for Evaluation-driven Text Public-Speaking Style Transfer}

\author{
Huashan Sun$^{*}$ \quad Yixiao Wu\thanks{~~Equal contribution} \quad \textbf{Yuhao Ye} \quad \textbf{Yizhe Yang} 
\\ \quad \textbf{Yinghao Li} \quad \textbf{Jiawei Li}   
 \quad \textbf{Yang Gao}\thanks{~~Corresponding author} \\
School of Computer Science and Technology, Beijing Institute of Technology\\ 
Beijing Engineering Research Center of High Volume Language Information Processing \\
        and Cloud Computing Applications\\
\texttt{\{hssun,yxwu,yhye,yizheyang,yhli,jwli,gyang\}@bit.edu.cn}\\
}

\begin{document}

\maketitle
\begin{abstract}
Language style is necessary for AI systems to understand and generate diverse human language accurately. However, previous text style transfer primarily focused on sentence-level data-driven approaches, limiting exploration of potential problems in large language models (LLMs) and the ability to meet complex application needs. To overcome these limitations, we introduce a novel task called \textbf{P}ublic-\textbf{S}peaking \textbf{S}tyle \textbf{T}ransfer (\textbf{PSST}), which aims to simulate humans to transform passage-level, official texts into a public-speaking style.
Grounded in the analysis of real-world data from a linguistic perspective, we decompose public-speaking style into key sub-styles to pose challenges and quantify the style modeling capability of LLMs. 
For such intricate text style transfer, we further propose a fine-grained evaluation framework to analyze the characteristics and identify the problems of stylized texts.
Comprehensive experiments suggest that current LLMs struggle to generate public speaking texts that align with human preferences, primarily due to excessive stylization and loss of semantic information\footnote{Model checkpoints and data resources are available at \href{https://github.com/shs910/PSST}{https://github.com/shs910/PSST}}. 
\end{abstract}

\section{Introduction}

Text Style Transfer (TST) is crucial in Natural Language Processing (NLP), focusing on modifying text style while retaining the original content's information~\citep{hu2022text,i:1-TST-survey}. By modeling complex human styles, including personality, habits, and mindset~\citep{i:1-TST-survey,geroda2023analysis}, AI models can further accurately understand and generate diverse human languages for user-centric applications such as role-playing~\citep{wang2023rolellm} and digital personas~\citep{Clarke1994TheDP,Digital_Persona}.

\begin{figure}
    \centering
    \includegraphics[scale=0.3]{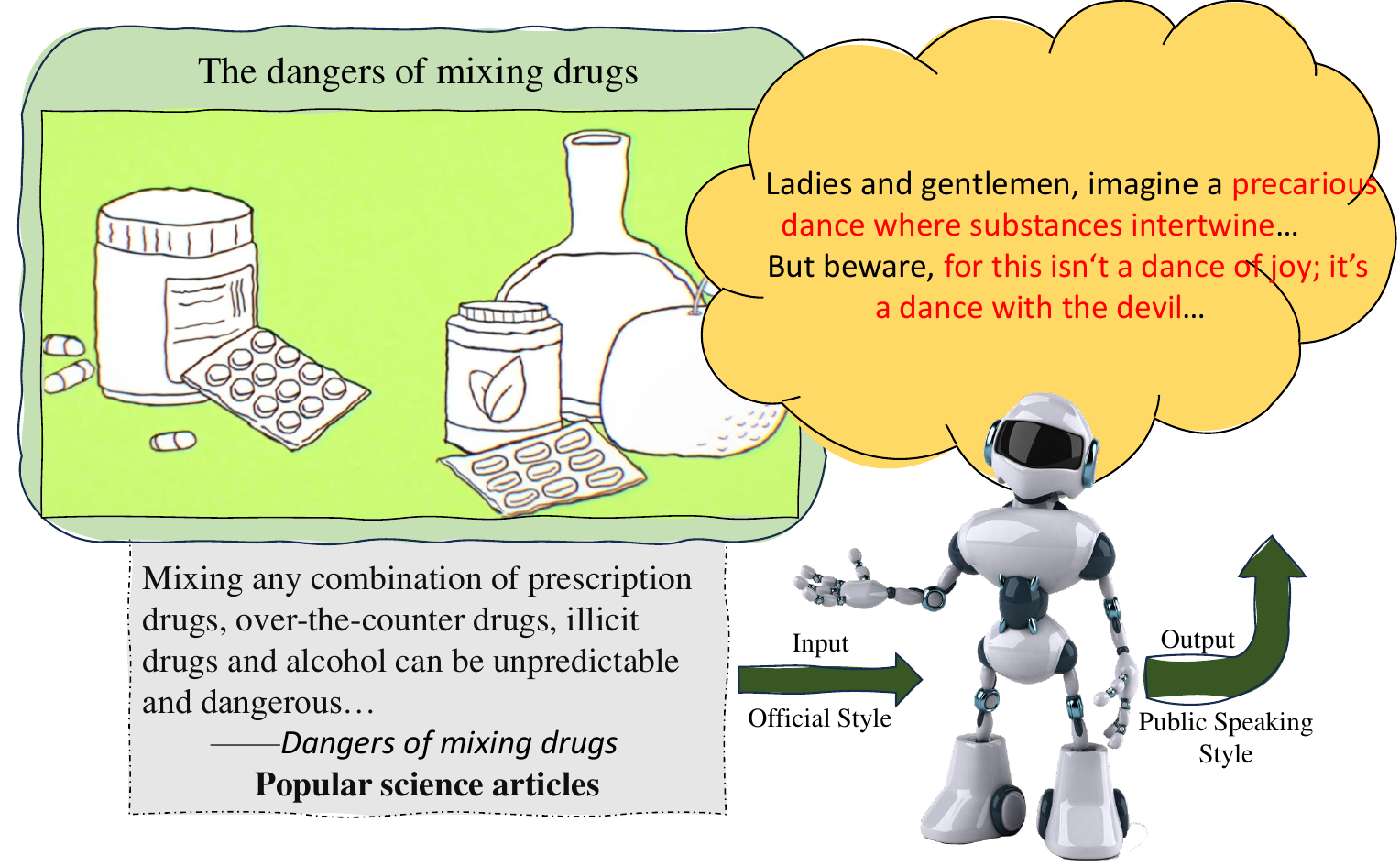}
    \caption{Illustration of Public-Speaking Style Transfer (PSST). An AI model is requested to present a written text, such as a popular science article, to audiences vividly and engagingly. The example generated by ChatGPT in the figure shows excessive stylization (highlighted in \textcolor{red}{red}).}
    \label{fig:PSST-example}
\end{figure}

Benefiting from their superior capabilities, large language models (LLMs)~\citep{i:1-GPT3,i:1-LLaMA-2,yang2023mindllm} can achieve high performance in traditional style transfer tasks~\citep{li-etal-2024-fundamental}. However, traditional style transfer is limited to sentence-level transformations and only a few data-driven styles~\citep{i:1-TST-survey} (e.g., formal-informal~\citep{sheikha2011generation}, polite-impolite~\citep{politeness}). This is insufficient for complex real-world applications, which require passage-level transformations and complex styles~\citep{i:1-TST-survey}.  In this paper, we introduce a novel task, called \textbf{P}ublic-\textbf{S}peaking \textbf{S}tyle \textbf{T}ransfer (\textbf{PSST}), which could further facilitate critical applications such as knowledge dissemination, public education, business promotion, etc. Specifically, PSST transforms formal, long paragraph texts into a public-speaking style, a form of language used by human beings to convey influential knowledge and ideas in public~\citep{Kedrowicz2016ShiftingRN,Beebe2005ThePS,DBLP:journals/tkde/GaoXHLWL20} (as shown in Figure~\ref{fig:PSST-example} and video demo in~\ref{sec:demo}). 




To investigate the language style modeling capabilities of LLMs in scenarios akin to PSST, we initially employ LLMs (eg. ChatGPT~\cite{chatGPT} and Llama 3-Instruct~\citep{llama3modelcard}) for PSST and identify three primary issues: (1) Over-stylization; (2) Uneven Style Strength Distribution; (3) Severe Semantic Degradation. To further quantify and enhance the language style modeling capabilities of LLMs, we propose a fine-grained evaluation framework tailored for complex long-text style transfer tasks such as PSST focusing on style strength and semantic preservation \citep{rao-tetreault-2018-dear, i:2-sentiment}, which enables a continuous, evaluation-driven approach to enhance the language style modeling capabilities of LLMs.  

For style strength evaluation, we propose two metrics: (1) \textit{passage-level style strength score} that offers a coarse-grained measure of the overall style strength, and (2) \textit{style strength distribution} that captures the distribution of style elements throughout the text. Using real data as a standard, we analyze the style modeling capability of LLMs based on the above metrics.
Specifically, we first gather official texts (eg. Encyclopedias) as source texts, and real public speaking data (eg. Ted Talks) as the target style dataset.
To accurately define and evaluate public-speaking style, we review linguistic literature on spoken public speeches~\citep{mccroskey2003principles,Beebe2005ThePS,Atkinson_1985,Halliday1989SpokenAW} and summarize four prominent sub-styles: \textit{orality}, \textit{interactivity}, \textit{vividness}, and \textit{emotionality}, supported by manual annotation.
Since we aim to quantify the style elements at various positions within the stylized text and LLMs have demonstrated strong capabilities in the sentence level TST~\citep{lai2023multidimensional}, for each sub-style, we generate a series of examples varying in style strength from a single sentence, and then we rank and score them with GPT3.5-Turbo\footnote{https://platform.openai.com/docs/models/gpt-3-5-turbo}. Then we train a small scorer (TinyLlama-1.1b~\citep{Zhang2024TinyLlamaAO}) to predict these scores. 

For semantic preservation evaluation, we propose a QA-based method. The accuracy changes of a QA model (eg. Llama3-8B-Instruct~\citep{llama3modelcard}) on texts before and after style transfer reflect the LLM's ability of semantic preservation. Specifically, we use GPT-4\footnote{https://openai.com/research/gpt-4} to generate high-quality QA pairs from the source text focusing on two dimensions: \textit{key information} and \textit{logical structure}~\citep{coopman2018public}. 


Moreover, comprehensive experiments show that prompt engineering can mitigate but not completely resolve the above problems, which highlights a gap between LLMs and humans in language style modeling, underscoring a substantial opportunity for improvement in this area. 

Our main contributions are as follows:
\begin{itemize}
\item We collect data and introduce a valuable and extensible task named Public-Speaking Style Transfer (Section~\ref{sec:PSST}).
We decompose the intricate text style into key sub-styles for more accurate style definition and evaluation (Section~\ref{Features_Analysis} and~\ref{sec:Key_Features_of_Public-Speaking_Style}).



\item We propose a fine-grained evaluation framework for PSST (Section~\ref{Evaluation_System}), which is extensible to incorporate additional sub-styles (e.g. personality), enabling an evaluation-driven approach to continuously analyze and enhance the language style modeling capabilities of LLMs.

\item We conduct a comprehensive evaluation of the performance of mainstream LLMs on PSST (Section~\ref{sec: evaluation_results}). Our analysis reveals that current LLMs often exhibit over-stylization, uneven style strength distribution, and severe semantic degradation.
\end{itemize}

\begin{table*}[ht]\small
    \centering
    \begin{tabular}{llll}
    \toprule
         \textbf{TST task}&\textbf{style}&\textbf{examples} &\textbf{features}\\
    \midrule
    \multirow{10}{*}{\makecell[c]{official\\$\leftrightarrow$\\p-s}}
    &official&GDOS is a modified version of WEDOS, which facilitates ...&\multirow{2}{*}{\makecell[l]{arousing\\interest}}\\
    &\makecell[l]{p-s}&\textbf{Have you heard of GDOS?} It's a modified version of ... \textbf{it helps you} ...&\\
    \cdashline{2-4}
    &official&\makecell[l]{Deeply concerned that the situation in Rwanda, which has resulted in\\ the death of many thousands of innocent civilians, including women and children}&\multirow{2}{*}{\makecell[l]{appropriate\\emotion}}\\
    &\makecell[l]{p-s}&\makecell[l]{\textbf{I'm really worried about} what's going on in Rwanda.\\\textbf{So many} innocent people, including women and children, have died.} &\\
    \cdashline{2-4}
    &official&\makecell[l]{Instead, they've become emboldened.}&\multirow{2}{*}{\makecell[l]{better \\vividness}}\\
    &\makecell[l]{p-s}&\makecell[l]{Instead, they have grown \textbf{stronger and more courageous.}} &\\
    \cdashline{2-4}
    &official&\makecell[l]{The Conference also agreed that the Bureau would keep the calendar under review}&\multirow{2}{*}{\makecell[l]{oral\\expression}}\\
    &\makecell[l]{p-s}&\makecell[l]{\textbf{So, uh}, the Conference also \textbf{agre-uh,} agreed that the Bureau would keep the\\calendarunder review, \textbf{y'know?}} &\\
    \midrule
    \multirow{4}{*}{\makecell[c]{formal\\$\leftrightarrow$\\informal}}&formal&He is very \textbf{attractive}&\multirow{4}{*}{\makecell[l]{Manuscript \\Form and \\Punctuation,\\Vocabulary}}\\
    &informal&he \textbf{iss wayyy hottt.}&\\
    \cdashline{2-3}
    &formal&\textbf{Yes}, but not for episode \textbf{IV}.&\\
    &informal&\textbf{yes}, except for episode \textbf{iv}.&\\
    \bottomrule
    \end{tabular}
    \caption{Comparison of PSST and formality TST: \textbf{a.} ``p-s'' means public-speaking style. \textbf{b.} Traditional TST primarily focuses on vocabulary-level adjustments, such as adhering to writing norms and word norms. In contrast, PSST involves audience engagement, including posing questions to generate interest and appeal to the audience.}
    \label{tab:task_comparise}
\end{table*}
\section{Related Work}
\label{sec:related_work}
\paragraph{Definition of Text Style}
Style, from a linguistic perspective, encompasses various elements that contribute to the conveyance of semantics, including word choice, sentence structure, and arrangement, all of which work together to establish the tone, imagery, and meaning in the text~\citep{mcdonald1985computational,hu2022text,Word_Matters}. In contrast, research on text style transfer (TST) takes a data-driven approach, defining style as attributes or labels based on style-specific corpora ~\citep{shen2017style,rao-tetreault-2018-dear}, which may affected by other attributes in dataset~\citep{i:1-TST-survey}. However, our approach is grounded in real data and explores key dimensions from the perspective of public-speaking linguistics. This definition alleviates the ambiguity and reduces the difficulty of data construction and evaluation. 
\paragraph{Evaluation of Text Style Transfer}
Firstly, fluency, a common objective in most natural language generation tasks, is often measured by the perplexity score (PPL)~\citep{yang2018unsupervised}. Secondly, to evaluate content preservation during the style transfer, metrics include BLEU~\citep{papineni2002bleu}, ROUGE~\citep{lin2004automatic}, BERTScore~\citep{zhang2019bertscore} are employed. Thirdly, the strength of style is an important dimension. Typically, a binary style classifier is first separately pretrained to predict the style label of input sentences~\citep{GAO2024103643,Thank_you_BART,two-stage}. This classifier is then used to estimate the style transfer accuracy.
Recently, ~\citet{lai2023multidimensional} demonstrates that ChatGPT~\citep{chatGPT} achieves competitive correlation with human judgments to serve as a multidimensional evaluator for sentence-level formal-informal style transfer.

To effectively evaluate long texts in PSST and facilitate detailed analysis, we propose a multi-dimensional fine-grained framework to assess the style strength and distribution of a long text and a QA-based method to capture differences in details and logic of the text before and after PSST.
\section{Public-Speaking Style Transfer}
\label{sec:PSST}
\begin{figure*}[ht]
    \centering
    \includegraphics[scale=0.5]{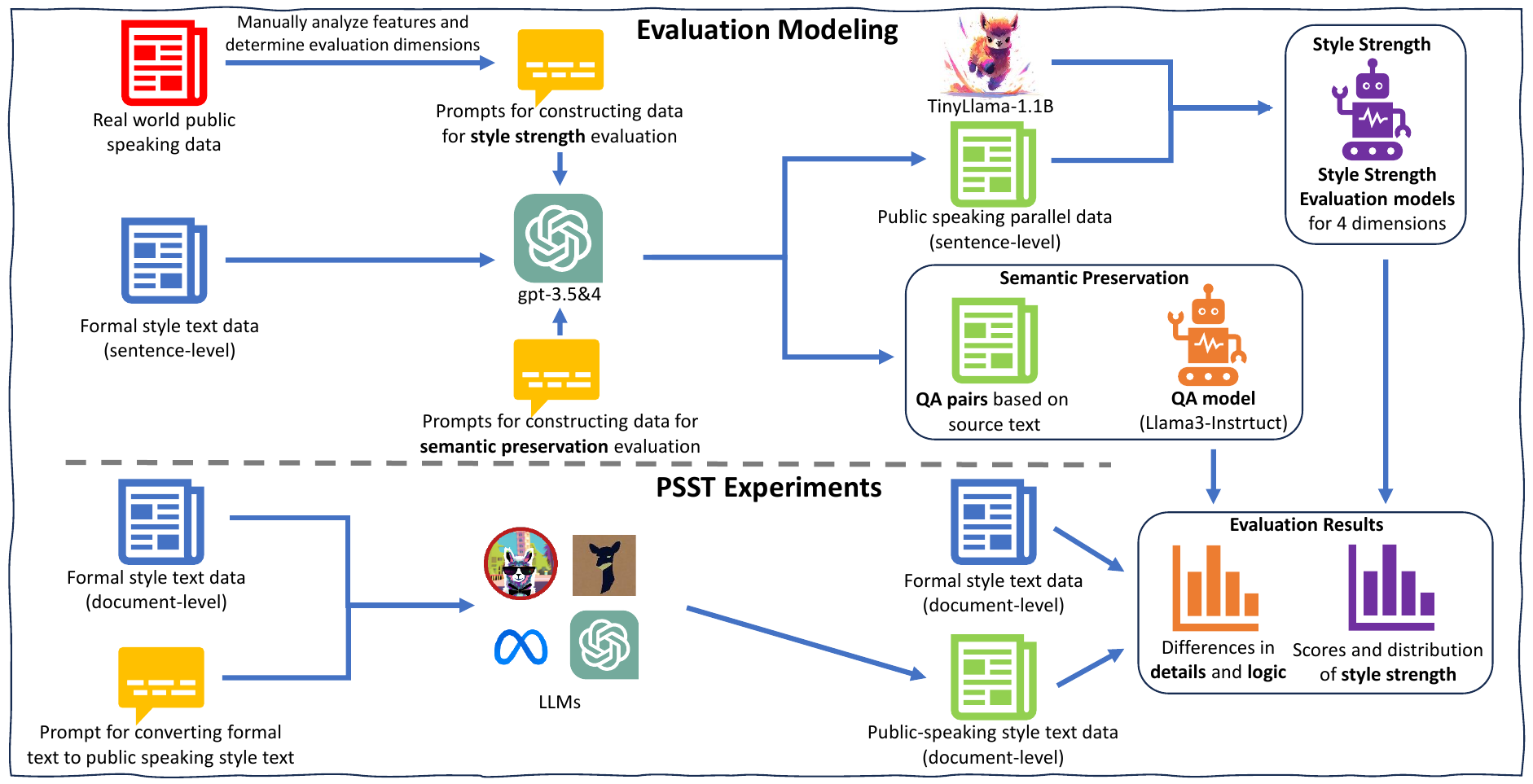}
    \caption{Pipeline of establishing the evaluation system of the PSST task and the experiment\&analysis of LLMs. \textbf{1.} The above depicts the process of establishing the evaluation system. Specifically, we begin with a comprehensive analysis of real-world data from a linguistic perspective and identify four key characteristics of public-speaking style. Subsequently, we employ GPT-3.5 to generate a sentence-level list-wise parallel corpus and train TinyLlama-1.1B as a scorer for each dimension. For semantic preservation, we utilize GPT-4 to generate QA pairs that focus on key information and logic in source texts. We then assess these pairs with a QA model applied to stylized text, using variations in model accuracy to evaluate semantic integrity. \textbf{2.} The bottom presents the experiment and analysis of the PSST task for the current LLMs.}
    \label{fig:PSST_pipeline}
\end{figure*}

In this section, we introduce the Text Public-Speaking Style Transfer (PSST) task. Initially, we provide a precise definition of PSST and describe the source dataset we constructed for this task. Furthermore, leveraging real-world public-speaking style data, we decompose the abstract public-speaking style into essential sub-styles, which are introduced in detail in Section~\ref{Features_Analysis}.
\subsection{Task Formulation}
\label{sec:task_formation}

The PSST task involves transforming official text style $a$ (e.g. news articles) into a more conversational and public-speaking-oriented language style $a'$, which can be formulated as:
\begin{equation}
    \begin{split}
        y(a')=&P(x(a)|a',[a'_1,\dots,a'_n])\\
    \end{split} 
\end{equation}
where $x(a)$ and $y(a')$ represent the official input text and the public-speaking style output, respectively. $[a'_1,\dots,a'_n]$ indicates additional conditions that can be introduced to enhance this task, including factors like the speaker's personality, the audience's specific preferences, and so forth.

As shown in Table~\ref{tab:task_comparise}, official-style texts typically employ specialized and complex vocabulary and sentence structures, conveying a serious and objective tone. In contrast, public-speaking style texts are more suitable for oral expression, characterized by simpler vocabulary and sentence patterns, and a more direct tone. Additionally, the public-speaking style features audience-oriented attributes, which we will elaborate on in Section~\ref{Features_Analysis}.
\subsection{Source Data}
\label{source_data}
Recall that our goal is for LLMs to emulate human public speaking to effectively convey knowledge. Therefore, we select three types of official texts with intensive knowledge as source texts: news articles\footnote{https://www.kaggle.com/datasets/clmentbisaillon/fake-and-real-news-dataset}, encyclopedias\footnote{https://huggingface.co/datasets/wikitext}, and research paper abstracts\footnote{https://www.kaggle.com/datasets/Cornell-University/arxiv}. For the target style dataset, we utilize data from real scenarios including TED talks\footnote{https://huggingface.co/datasets/iwslt2017/viewer/iwslt2017-
en-zh}, political speeches\footnote{https://www.americanrhetoric.com/}, academic presentations\footnote{https://iwslt.org/2023/multilingual}, and educational lectures\footnote{https://www.webpages.uidaho.edu/psyc390/index.htm}.


To ensure the reliability of our evaluation and comprehensively assess the model's capability for style transfer, we further select the source and target datasets by filtering based on token counts ($400\pm100$, $800\pm200$, $1200\pm200$) and ensuring comparable lengths between the two. The final dataset used for PSST is shown in Table~\ref{tab:final_dataset_statistics}. See Appendix~\ref{Dataset} for a detailed description and discussion.


\subsection{Prior Fine-grained Analysis}
\label{Features_Analysis}

To achieve a more precise style definition and evaluation, we deconstruct the public-speaking style into several key sub-styles. We initially review research papers focused on oral public speaking~\citep{mccroskey2003principles,Beebe2005ThePS,Atkinson_1985,Halliday1989SpokenAW} and summarize the following key candidate features: (1) \textit{Interactivity}, (2)  \textit{Emotionality}, (3)  \textit{Filler Words}, (4)  \textit{Vividness}, (5)  \textit{Ambiguity}, (6)  \textit{Abbreviations}, (7)  \textit{Informal Lexicon}. See Appendix~\ref{sec:human_annotation_for_Prior_Fine-grained_Analysis} for detailed descriptions. Given the diversity of public speaking scenarios, we further employ rigorous manual annotation to identify the prominent features.



Specifically, we randomly sample 300 sentences from the target style dataset mentioned in Section~\ref{source_data}, which 3 annotators then annotate based on two evaluation approaches. In the multi-label approach, we select labels that signify the public-speaking style features present in a specific instance. In contrast, the best-one approach selects the most salient label that embodies the public-speaking style of the sentence. Details can be found in Appendix~\ref{sec:human_annotation_for_Prior_Fine-grained_Analysis}. Note that we further provide some detailed guidelines to reduce the abstraction of sub-styles for a more accurate annotation.

\begin{figure}[!ht]
\begin{center}
\includegraphics[width=1\linewidth]{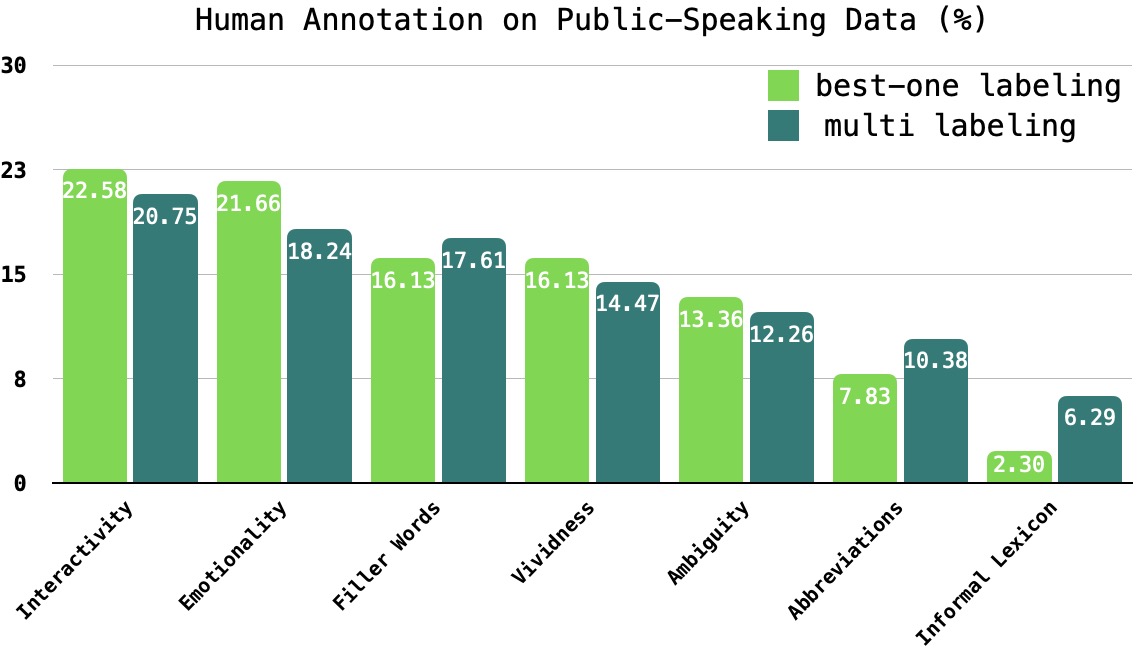} 
\caption{Human annotation on features of real public speaking data (Inner-Annotator Agreements: Krippendorff's $\alpha$ = 0.7773)\protect\footnotemark. "Interactivity", "Emotionality", "Filler words" and "vividness" are notable features.}
\label{fig:oral_label_res}
\end{center}
\end{figure}

\footnotetext{All Krippendorff's $\alpha$ coefficient in this paper are computed by the \texttt{alpha} function of Python’s \texttt{krippendorff} library~\citep{castro-2017-fast-krippendorff}}


The statistics in Figure~\ref{fig:oral_label_res} indicate that \textit{interactivity}, \textit{Emotionality}, \textit{Filler Words}, \textit{Vividness} are prevalent in the real dataset (indicated by multi labeling) and exhibit a pronounced tendency to emphasize public-speaking style (indicated by best-one labeling). Further, we merge \textit{Filler Words}, \textit{Ambiguity}, \textit{Abbreviations}, and  \textit{Informal Lexicon} into one feature \textit{Orality} as the following two reasons: 
(1) We find these features frequently co-occur, particularly in descriptions where multiple sentences of simple structure are employed alongside filler words to facilitate natural transitions. (2) These features collectively contribute to the oral nature of the speech~\citep{Halliday1989SpokenAW} but do not individually stand out as strongly as the primary features. 




\subsection{Key Features of Public-Speaking Style}
\label{sec:Key_Features_of_Public-Speaking_Style}
Based on the above analysis, we set out the following four dimensions to categorize the characteristics of public-speaking style and emphasize the distinctions among these categories. 
\begin{itemize}
\item{\textbf{Interactivity:} Interactivity in public speaking refers to the speaker engaging with the audience through various means such as posing thought-provoking, facilitating personal reflection, and crafting intriguing hypothetical scenarios (Table~\ref{tab:task_comparise} example 1)}.
\item{\textbf{Emotionality:} Public speaking contains the speaker's appropriate views and attitudes on specific events to reflect the speaker's emotional tendencies and inner thoughts (Table~\ref{tab:task_comparise} example 2)}.
\item{\textbf{Vividness:} 
In public speaking, speakers should present information in a lively, easy-to-understand way, such as using analogies and metaphors to make complex ideas more accessible and engaging (Table~\ref{tab:task_comparise} example 3)}.

\item{\textbf{Orality:}  The public-speaking style texts should align with oral communication norms, incorporating appropriate filler words, simple sentence structures, and suitable word choices (Table~\ref{tab:task_comparise} example 4)}
\end{itemize}.

We further offer some guidelines to mitigate the abstraction of sub-styles during human evaluation and data generation processes (eg. Figure\ref{fig:human_annotation_correlation-3-2}).
\section{Fine-grained Evaluation System}
\label{Evaluation_System}
The evaluation of PSST primarily focuses on \textbf{style strength} and \textbf{semantic preservation}. 
For style strength evaluation of long text, we propose a fine-grained evaluation framework in Section~\ref{sec:style_strength_eval}. For semantic preservation, we introduce a QA-based approach in Section~\ref{sec:semantic_preservation_eval}, which focuses on key details and logical structure.
It is noteworthy that the evaluation framework we propose is scalable. It is applicable across various scenarios, particularly those involving complex and abstract styles.
\subsection{Style Strength Evaluation}
\label{sec:style_strength_eval}
To evaluate the style strength of long texts in the PSST task, we propose a multi-dimensional, fine-grained evaluation method. Specifically, we generate a series of examples with varying style strengths from a single sentence, rank and score them using \texttt{gpt-3.5-turbo}, and then train a small scorer (TinyLlama-1.1b~\citep{Zhang2024TinyLlamaAO}) to predict these scores. This approach is appropriate for three main reasons:
\textbf{(1)} Early experiments suggest that using LLMs directly as evaluators may not be feasible (Section~\ref{sec:correlation}).
\textbf{(2)} Relevant studies show that LLMs possess strong stylization and evaluation capabilities in TST tasks at the sentence level~\citep{lai2023multidimensional}.
\textbf{(3)} For long texts, a simple overall style strength score does not facilitate detailed analysis. We believe that a similar style of two long texts should be reflected in the overall style strength and the stylization of specific positions within the documents, such as the beginning and the end.

\subsubsection{Fine-grained Evaluation Modeling}
To minimize evaluation costs and enhance the applicability of our evaluation framework, we distill the style evaluation capabilities of \texttt{gpt-3.5-turbo} into smaller models.

We utilize \texttt{gpt-3.5-turbo} to generate five sentences from an official sentence, ensuring consistent semantics while progressively increasing the style strength. Each sentence is scored on a scale from 1 to 5. To maintain precise ordering of stylistic strength and corresponding scores, we employ \texttt{gpt-3.5-turbo} as both the generator and evaluator within a single prompt.
Please refer to Appendix~\ref{sec:prompt_for_generate_data} for detailed prompts.

Then, we fine-tune TinyLlama-1.1b~\citep{Zhang2024TinyLlamaAO} as the sentence-level scorer named $\text{\textbf{EvalModel-1.1B}}_{\text{\textbf{gpt-3.5}}}$ for each dimension. Please refer to Appendix~\ref{Dataset_Training_Details} for more detailed data statistics and training implementation.
\subsubsection{Text-Level Style Evaluation}
In our evaluation framework, given a document $D$ we first use \texttt{stanza}\footnote{https://stanfordnlp.github.io/stanza/} to segment long text into sentences $S=\{s_1,s_2,\dots,s_M\}$. Then, we score the sentences using $\text{\textbf{EvalModel-1.1B}}_{\text{\textbf{gpt-3.5}}}$ in an $N$-gram manner, which means that we combine each of the $n$ sentences and score them, for example, $[s_1,s_2,s_3],[s_2,s_3,s_4],\dots,[s_{M-2},s_{M-1},s_{M}]$ where $n=3$.
The corresponding score sequence is denoted as $Score_{seq-n}$, and in our experiment we set $n\in\{1,2,3,4\}$.

For text-level evaluation, we implement two metrics to assess different LLMs. The first metric, named \textit{Text-Level Style Score}, can be used to make coarse-grained comparisons of different texts. 
\begin{equation}
    Score_{text}=\frac{1}{4}\sum_{n\in\{1,2,3,4\}} \texttt{mean}(Score_{seq-n})
\end{equation}
The second metric, named \textit{Style Score Distribution}, can be used for fine-grained comparison of style strength in different positions of texts. Specifically, given a document $D$ and its $N$-gram score results $\{Score_{seq-n}\}$, we chunk different $N$-gram sequences into $K$ segments (we set $K=5$), and then average them by position to obtain the style strength distribution:
\begin{equation}
    Score_{dist}=[s_1,s_2,\dots,s_K]
\end{equation}

\subsubsection{Correlation}
\label{sec:correlation}
For computing correlation, we collect a dataset comprising 100 sets of samples, each containing four texts generated by different models. We then ask three evaluators to rank each set of texts by style strength to establish the ground truth with the questionnaire shown in Figure~\ref{fig:human_annotation_correlation}. $\text{\textbf{EvalModel-1.1B}}_{\text{\textbf{gpt-3.5}}}$ and \textbf{Llama 2-Chat-70B}~\footnote{https://huggingface.co/meta-llama/Llama-2-70b-chat-hf}  are used to score and rank candidate texts. The prompt used by Llama 2-Chat-70B is shown in Figure~\ref{fig:prompt_for_evaluation}, which we refer to the method in ~\citet{lai2023multidimensional}.



\begin{table}[ht]
    \centering
    \begin{tabular}{lcc}
    \toprule
\textbf{EvalModel}&\makecell[c]{Llama 2-Chat\\(70B)}&\makecell[c]{Ours\\(1.1B)}\\
    \midrule
    Emotionality&56.98&76.29\\
    Interactivity&50.05&81.54\\
    Vividness&51.11&84.50\\
    Orality&13.98&72.94\\
    \hline
    Average&43.03&78.82\\
    \bottomrule
    \end{tabular}
    \caption{\textbf{Spearman’s} $\rho$ between different evaluation models and human evaluation (Inner-Annotator Agreements: Krippendorff's $\alpha$ = 0.8163). \textbf{a.} Ours(1.1B), exactly the $\texttt{EvalModel-1.1B}_{\texttt{gpt-3.5}}$, denotes style-strength evaluation model trained utilizing datasets derived from GPT-3.5 for each dimension.}
    \label{tab:correlation_1}
\end{table}
Results shown in Table~\ref{tab:correlation_1} indicate that the fine-grained evaluation method proposed in this paper outperforms the LLM-based evaluation method. We hypothesize that the sentence-level assessment method provides a more granular evaluation. For illustrative instances, please refer to Appendix~\ref{Bad_case_of_Llama2_in_Evaluation}.

\begin{figure*}[t]
    \centering
    \includegraphics[scale=0.35]{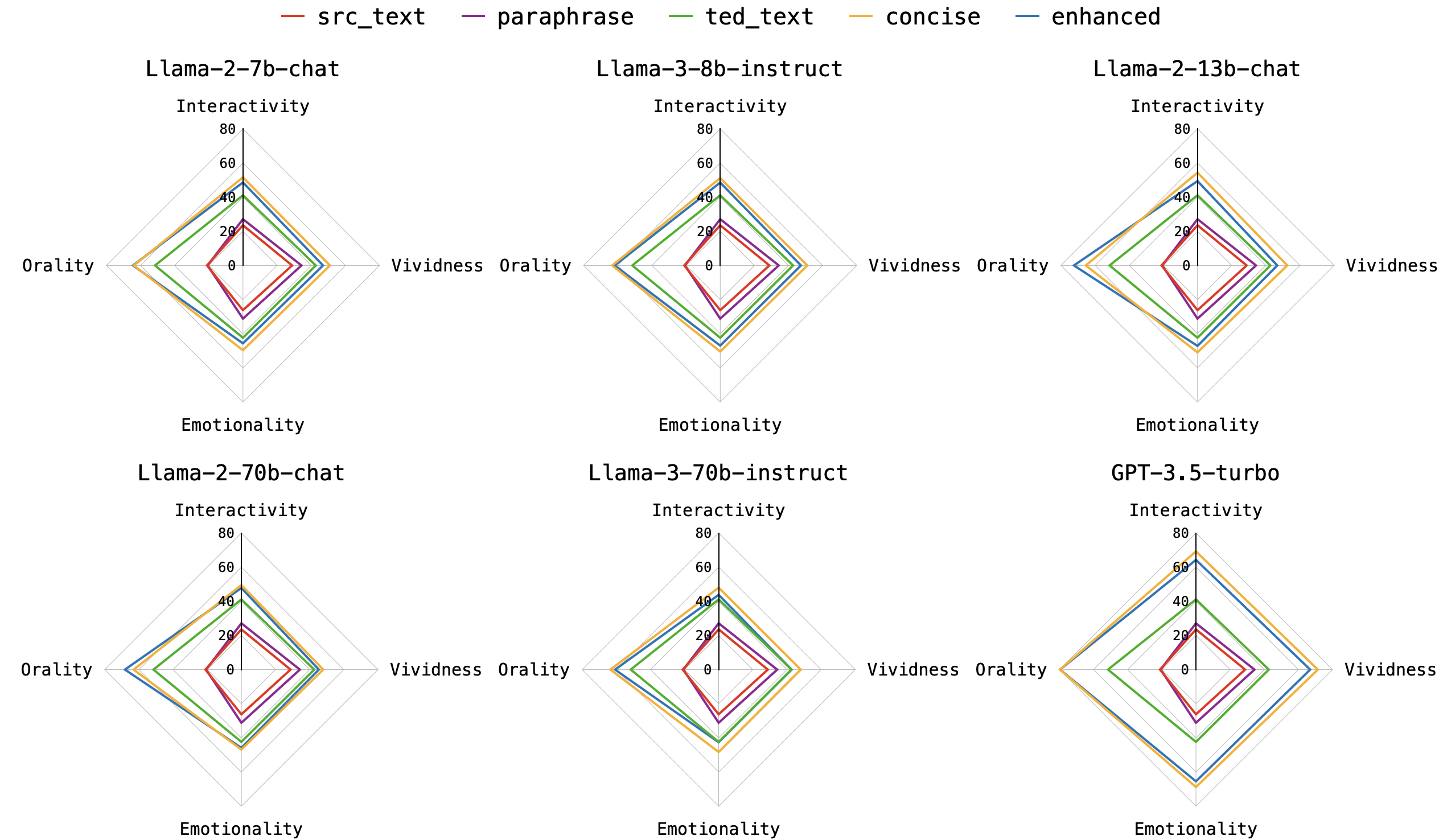}
    \caption{Radar plot of text-level style strength of passages transferred by different LLMs.($800\pm200$ tokens).}
    \label{fig:radar-800}
\end{figure*}

\subsection{Semantic Preservation}
\label{sec:semantic_preservation_eval}
For semantic preservation, we propose a QA-based approach, which we believe is better suited to public-speaking real-world scenarios—the style transfer model as a speaker and the QA model as a listener.
\subsubsection{QA-based Evaluation Method}
To guarantee the quality of QA pairs, we use \texttt{gpt-4} to generate multiple-choice questions that capture the details and logical relationships inherent in the original text, which are important elements of public speaking~\citep{coopman2018public}. We then test the QA model (Llama-3-8B-Instruct) on the text before and after style transfer. Variations in the model's accuracy indicate the degree of semantic preservation. Additionally, we use text paraphrased by \texttt{gpt-3.5-turbo} as a baseline for comparison. Specifically, we evaluate content preservation from the following two aspects: 
\begin{enumerate}
    \item[(1)] \textbf{Key Information}: This evaluates whether the essential information, facts, and details from the original text are preserved. Consequently, these questions should be answerable directly from the text, without requiring reasoning or external knowledge, and should include a variety of question types (e.g., "What," "Who," "When," "Where," "How").
    \item[(2)] \textbf{Logical Structure}: This examines whether the emotions, logical relationships and role relationships inherent in the original text are preserved. Consequently, the questions must delve into the specifics of the content and reasoning, such as identifying the speaker or audience, understanding the emotions, and analyzing the logical relationships within the text.
\end{enumerate}
The above two multiple-choice questionnaires consist of 10 question-options pairs and correct answers for each source text. Detailed prompts can be found in Appendix~\ref{sec:prompt_for_generate_QA_pairs}.

\begin{table}[]
    \centering
    \begin{tabular}{lcc}
    \toprule
         \textbf{Metric}&\textbf{Spearman's $\rho$ }&\textbf{Krippendorff's $\alpha$}\\
    \midrule
         BLUERT&60.00&60.67 \\
         BertScore&66.67&67.22 \\
         QA-based&\textbf{75.00}&\textbf{75.42} \\
    \bottomrule
    \end{tabular}
    \caption{Correlation between different semantic preservation evaluation methods and human evaluation. (inter-annotator consistency: Krippendorff’s $\alpha$ = 0.7693)}
    \label{tab:Results_of_Correlation_Semantic_preservation_1}
\end{table}
\subsubsection{Correlation}
We conduct three manual evaluation experiments to verify the effectiveness of the QA-based method.

Firstly, as shown in Table~\ref{tab:Results_of_Correlation_Semantic_preservation_1}, our QA-based method exhibits a high degree of correlation with human evaluation, outperforming BLUERT~\citep{sellam-etal-2020-bleurt} and BertScore~\citep{zhang2019bertscore} (which also have a limited processable context length of 512). Secondly, we manually examine whether the QA-pairs detect semantic loss in texts before and after style transfer. The results show that \textbf{87/100} of the QA pairs successfully identify the corresponding message missing in the stylized texts. Thirdly, we test the QA model on QA-pairs generated based on stylized text. The accuracy of the QA model’s responses was \textbf{98\%}, which indicates that the QA model has a robust capability for handling Public-Speaking Style texts. 

The experiments above demonstrate that the decreased accuracy of the QA model is indeed attributed to the semantic loss occurring during the style transfer process. Please refer to Appendix~\ref{sec:annotation_for_Semantic_Preservation} for more details. 

\section{Public Speaking Ability Evaluation}
\label{sec:exps}
\begin{figure*}[ht]
    \centering
    \includegraphics[scale=0.35]{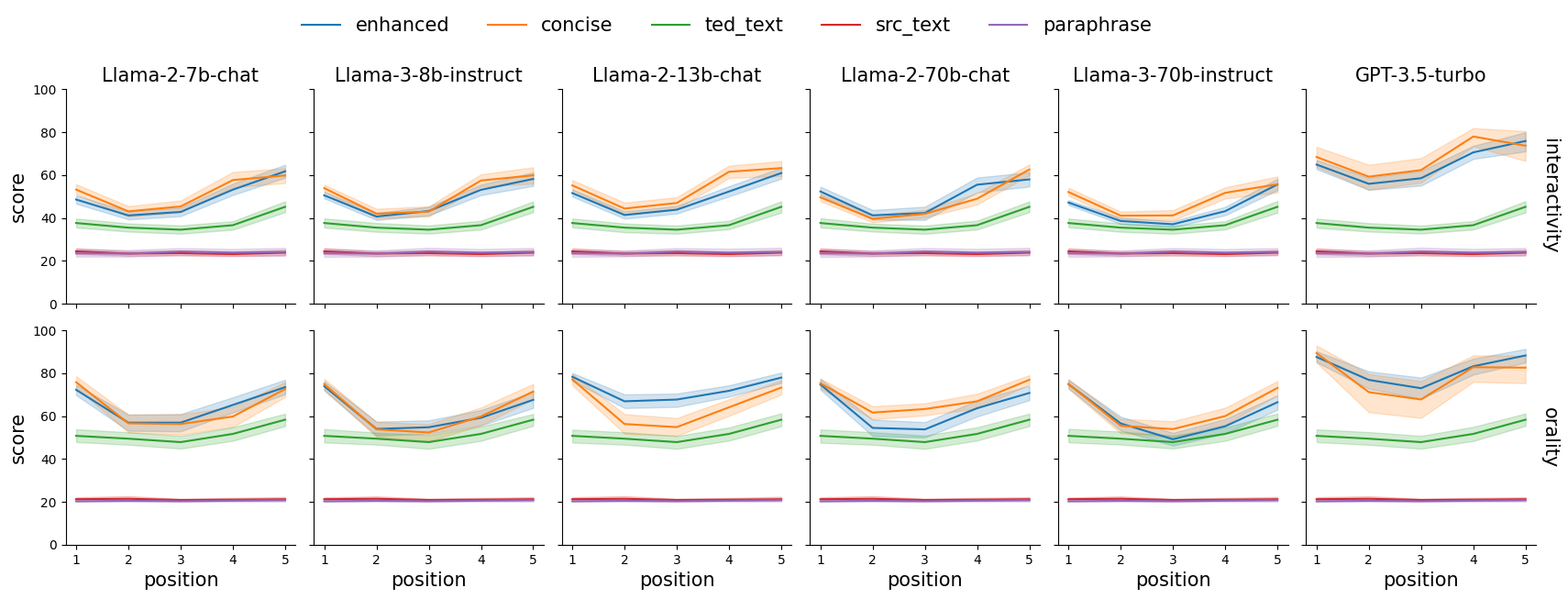}
    \caption{Style strength distribution of passages ($800\pm200$ tokens) transferred by different LLMs in \textit{Interactivity} and \textit{Orality}. }
    \label{fig:style_strength_distribution}
\end{figure*}
\subsection{Experiment Settings}
\label{sec:overall_setups}
\paragraph{Baselines} We assessed the public-speaking style modeling capability of the LLM by comparing it with the following text. (1) \textbf{\textit{src\_text}}, the official texts mentioned in Section~\ref{source_data}, characterized by an official style and complete information, serving as the upper bound for semantic preservation. (2) \textbf{\textit{paraphrase}}, texts paraphrased by \texttt{gpt-3.5-turbo} based on the source text, reflecting inherent changes in the language processing of LLMs. (3) \textbf{\textit{ted\_text}},
the target style text in real public speaking scenario. Note that \textbf{src\_text} and \textbf{ted\_text} are not parallel. Both \textbf{src\_text} and \textbf{paraphrase} have low public-speaking style strength but high semantic preservation scores. The closer the style strength of the generated texts is to real-world data and the lower their semantic loss, the stronger the model's capability in language style modeling.

\paragraph{Prompts} (1) \textbf{\textit{concise prompt}}, a brief instruction requiring models to simulate lively and engaging public speaking through style transfer. (2) \textbf{\textit{enhanced prompt}}, detailed guidance on the dimensions of stylization, cautioning against excessive stylization, and emphasizing the importance of semantic consistency. Five prompts are created for each type shown in Figure~\ref{fig:prompt4LLMs2PSST}.

\paragraph{Models} We employ Llama 2-Chat~\citep{i:1-LLaMA-2}, Llama 3-Instruct~\citep{llama3modelcard} and \texttt{gpt-3.5-turbo} to generate public-speaking style texts based on \textbf{src\_text} and \textbf{prompts}. We believe these models capture the current state of LLMs and can help explore the gap between LLMs and human abilities in complex, abstract language style modeling.
\subsection{Evaluation Results and Analysis}
\label{sec: evaluation_results}
Figures~\ref{fig:radar-800} and \ref{fig:style_strength_distribution} illustrate the overall text-level style strength score and the fine-grained style strength distribution for texts of 800 tokens, respectively. We adjust the scale of style strength score from 1–5 to 20–100 for better visualization. The results of the semantic preservation evaluation are presented in Figure \ref{fig:semantic-800}. Results for $400\pm100$ and $1200\pm200$ tokens are shown in Appendix~\ref{sec:additional_exps}, which have a similar trend.
\paragraph{Over-stylization}
Firstly, the transferred texts generated by different models using concise prompts and enhanced prompts (which discourage over-stylization) exhibit stronger style strength at the text level in each dimension compared to TED-Talks data, particularly in \textit{Orality}. Secondly, for GPT-3.5, known for its strong instruction-following ability, the style strength in each dimension is excessively higher than in real-world scenarios. For example, as shown in Figure~\ref{fig:PSST-example} and Case 2 in Figure~\ref{case_2}, there are too many style elements. Furthermore, the stylization of the underlined parts in Figure~\ref{case_2} seems to be intentionally exaggerated and appears unnatural, diverging from human preferences that do not align with human preferences. We assume that the model may be unable to truly understand the modeling process of language style in real public speaking scenarios and instead executes instructions mechanically. Thirdly, the style strength of LLaMA3s is closer to the real style strength than that of LLaMA2s. Moreover, Llama-3-70-Instruct performs best, with style strength closer to real-world norms in all four dimensions, particularly after the use of enhanced instructions. However, its style strength in \textit{Orality} remains relatively high.
\paragraph{Uneven Style Strength Distribution}
The results comparing style strength distribution across different positions within texts are illustrated in Figure~\ref{fig:style_strength_distribution}.  Specifically, in terms of interactivity and orality, the style distribution in real-world scenarios is more uniform, with slight increases in stylization at the opening and closing segments.  This pattern aligns with the conventional use of introductory and concluding phrases in real-world settings.  However, for LLMs, the distribution exhibits a ``U-shaped'' pattern, indicating excessive stylization at the text's beginning and end.  For instance, as depicted in Figure~\ref{fig:style_strength_distribution}, compared to real-world texts, LLMs such as GPT-3.5, Llama-3-70B-Instruct, and Llama-2-Chat-70B demonstrate pronounced stylization at the beginning and end, but less so in the middle sections (Case in Figure~\ref{case_0}).  This disparity suggests that LLMs diverge from human-like style modeling, potentially due to a model-inherent tendency to focus on the beginning and end of long texts, or possibly indicative of the model's ``lazy'' behavior.
\begin{figure}[ht]
    \centering
    \includegraphics[scale=0.18]{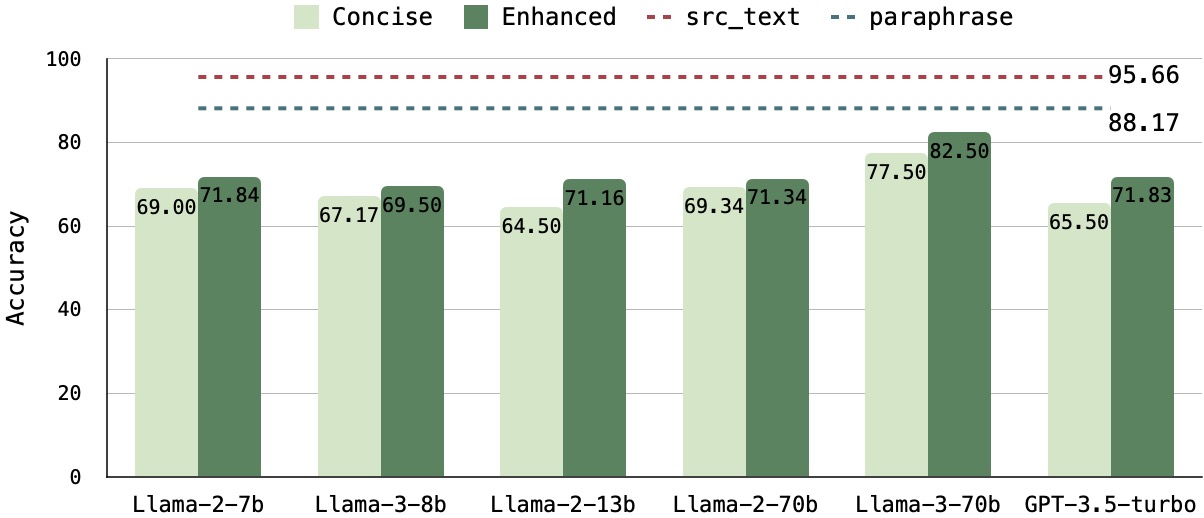}
    \caption{Semantic preservation based on QA (800$\pm$200 tokens).}
    \label{fig:semantic-800}
\end{figure}
\paragraph{Severe Semantic Degradation}
The results shown in Figure~\ref{fig:semantic-800} reveal significant semantic loss during style transfer. The high accuracy observed in \textit{src\_text} and \textit{paraphrase} confirms that the QA model can effectively respond to questions based on non-stylized texts. Moreover, our questions, designed to target essential information, are relatively straightforward and can be answered provided no information is lost. Therefore, the observed decline in accuracy with stylized texts further underscores the challenges LLMs face in semantic preservation. As shown in Figure~\ref{case_3}, the stylized text omits detailed descriptions of key steps, replacing them with generalized and vague statements, which is unfavorable, particularly in public education contexts such as popular science communication. In Figure~\ref{case_4}, the stylized text lacks a temporal reference for the specific event, which impairs the comprehensive representation of the event, particularly in the dissemination of critical information, such as news reporting. Notably, models exhibiting stronger stylization (e.g., LLaMA-3-70B-Instruct vs. GPT-3.5-Turbo) tend to obscure key information more. This is intuitive, as excessive stylization reduces the clarity of semantic content, thus hindering message comprehension for audiences, even if the essential points remain present. Additionally, while enhanced prompts focus on semantic preservation and achieve higher accuracy, they are still less than satisfactory.

\section{Conclusion}

In this paper, we introduce public-speaking style transfer, which requires LLMs to transform a formal and long text into a public-speaking style.
By analyzing real-world data from a linguistic perspective, decompose the intricate text style into key sub-styles. We propose a fine-grained evaluation framework that enables an evaluation-driven approach to continuously assess and enhance the language style modeling capabilities of LLMs. For style strength evaluation, we distill the LLM's ability to assess the style elements of sentences into a smaller model for each sub-style. We introduce two metrics: \textit{passage-level style strength score} and \textit{style strength distribution}, allowing for a detailed analysis of long stylized texts. For semantic preservation, we propose a QA-based method that focuses on key information and logical structure. By observing changes in the QA model's accuracy, we identify and analyze the LLM's ability to retain semantics. Our experiments reveal that current LLMs exhibit issues such as over-stylization, uneven style strength distribution, and severe semantic degradation in long and complicated language style modeling. These findings highlight the shortcomings in LLMs' language style modeling capabilities and underscore the substantial room for improvement.

\section*{Limitations}
\subsection*{Improvement of Evaluation Methods} 
\paragraph{Effective Evaluation of Planing in Public-Speaking} Style modeling in public speaking might involve elements related to planning. In this regard, incorporating insights from psychological language modeling~\citep{Psycholinguistics,Computational_Psycholinguistics} could help design more effective evaluation methods. Furthermore, we could develop a dynamic assessment method, such as incorporating "role-playing," to adaptively adjust evaluation results across different dimensions during the interactive process between various roles. This dynamic simulation of human activities would be more effective in assessing complex human abilities, such as "planning." 

\paragraph{Punishment Mechanism} The current evaluation system overlooks the "punishment rule." For instance, if a model employs excessively exaggerated descriptions in its speech, it may receive a higher score in the "vividness" or "orality" dimension. However, such descriptions may not align with our expectations. 

\subsection*{Domain and Style Extensions}
\paragraph{Limited Dataset Domain} While the current dataset has limited scope, our future plans involve expanding it to include more meaningful domains, such as health, sports, business, and education. 
\paragraph{Limited Substyle Analysis} This paper examines only four sub-style dimensions, omitting an analysis of more intricate scenarios. By employing a decomposition approach, we can continuously introduce additional styles, such as the daily spoken speech presentations revealing personalities. This method allows us to explore a wide range of stylistic variations and experiment with diverse ways of delivering oral content, enhancing the adaptability and creativity of our spoken language generation.


\subsection*{Token Length and Variety of Tested Models}
\paragraph{Limited Token Length}  Given the substantial semantic losses observed in current LLMs, we advocate for studying their complex style modeling capabilities within bounds that are reasonable given their current performance levels. 

\paragraph{Limited Variety of Models Tested} We primarily focused on testing mainstream, widely used, and capable LLMs (such as LLaMA and GPT-3.5). We believe these models are representative of the current landscape of LLMs and can effectively explore the gap between current LLMs and human abilities in complex, abstract language style modeling.
\\
\\
In future work, we plan to investigate longer texts and a broader range of models to further validate and demonstrate the generalizability of our findings.
\section*{Ethics Statement}
Here we discuss the primary ethical considerations of PSST.
\paragraph{Intellectual Property Protection} The utilized data is publicly available, permitting the reproduction, utilization, and modification of its content.
\paragraph{Content and Impact} 
The PSST task mandates the model to rephrase the text, potentially leading to the generation of inaccurate information, despite our efforts to maintain consistency with the original text through prompts. Simultaneously, we note that the transformation of speech style increases the likelihood of the model producing a more "inflammatory" language style. The observed outcomes do not manifest violence, discrimination, or other related issues. However, these are aspects that warrant additional scrutiny and should be addressed in future research.

\section*{Acknowledgements}
We thank Xinyue Liang and Shunyu Wang for their support in human annotation for this study. 

\bibliographystyle{acl_natbib}
\bibliography{emnlp2023}

\begin{thebibliography}{41}
\expandafter\ifx\csname natexlab\endcsname\relax\def\natexlab#1{#1}\fi

\bibitem[{AI@Meta(2024)}]{llama3modelcard}
AI@Meta. 2024.
\newblock \href {https://github.com/meta-llama/llama3/blob/main/MODEL_CARD.md} {Llama 3 model card}.

\bibitem[{Atkinson(1985)}]{Atkinson_1985}
J.~Maxwell Atkinson. 1985.
\newblock \emph{Public speaking and audience responses: some techniques for inviting applause}, Studies in Emotion and Social Interaction, page 370–410. Cambridge University Press.

\bibitem[{Beebe and Beebe(2005)}]{Beebe2005ThePS}
Steven~A. Beebe and Susan~J. Beebe. 2005.
\newblock \href {https://api.semanticscholar.org/CorpusID:141740745} {The public speaking handbook}.

\bibitem[{Brown et~al.(2020)Brown, Mann, Ryder, Subbiah, Kaplan, Dhariwal, Neelakantan, Shyam, Sastry, Askell, Agarwal, Herbert-Voss, Krueger, Henighan, Child, Ramesh, Ziegler, Wu, Winter, Hesse, Chen, Sigler, Litwin, Gray, Chess, Clark, Berner, McCandlish, Radford, Sutskever, and Amodei}]{i:1-GPT3}
Tom~B. Brown, Benjamin Mann, Nick Ryder, Melanie Subbiah, Jared Kaplan, Prafulla Dhariwal, Arvind Neelakantan, Pranav Shyam, Girish Sastry, Amanda Askell, Sandhini Agarwal, Ariel Herbert-Voss, Gretchen Krueger, Tom Henighan, Rewon Child, Aditya Ramesh, Daniel~M. Ziegler, Jeffrey Wu, Clemens Winter, Christopher Hesse, Mark Chen, Eric Sigler, Mateusz Litwin, Scott Gray, Benjamin Chess, Jack Clark, Christopher Berner, Sam McCandlish, Alec Radford, Ilya Sutskever, and Dario Amodei. 2020.
\newblock \href {http://arxiv.org/abs/2005.14165} {Language models are few-shot learners}.

\bibitem[{Castro(2017)}]{castro-2017-fast-krippendorff}
Santiago Castro. 2017.
\newblock Fast {K}rippendorff: Fast computation of {K}rippendorff's alpha agreement measure.
\newblock \url{https://github.com/pln-fing-udelar/fast-krippendorff}.

\bibitem[{Clarke(1994)}]{Clarke1994TheDP}
Roger~A. Clarke. 1994.
\newblock \href {https://api.semanticscholar.org/CorpusID:11607575} {The digital persona and its application to data surveillance}.
\newblock \emph{Inf. Soc.}, 10:77--92.

\bibitem[{Coopman and Lull(2018)}]{coopman2018public}
Stephanie~J Coopman and James Lull. 2018.
\newblock \emph{Public speaking: The evolving art}.
\newblock Cengage.

\bibitem[{Crocker and Brouwer(2023)}]{Computational_Psycholinguistics}
Matthew~W. Crocker and Harm Brouwer. 2023.
\newblock \href {https://doi.org/10.1017/9781108755610.032} {\emph{Computational Psycholinguistics}}, 2 edition, Cambridge Handbooks in Psychology, page 890–920. Cambridge University Press.

\bibitem[{Gao et~al.(2024)Gao, Liu, Yang, and Wang}]{GAO2024103643}
Yang Gao, Qianhui Liu, Yizhe Yang, and Ke~Wang. 2024.
\newblock \href {https://doi.org/10.1016/J.IPM.2024.103643} {Latent representation discretization for unsupervised text style generation}.
\newblock \emph{Inf. Process. Manag.}, 61(2):103643.

\bibitem[{Gao et~al.(2020)Gao, Xu, Huang, Liu, Wei, and Liu}]{DBLP:journals/tkde/GaoXHLWL20}
Yang Gao, Yue Xu, Heyan Huang, Qian Liu, Linjing Wei, and Luyang Liu. 2020.
\newblock \href {https://doi.org/10.1109/TKDE.2019.2892430} {Jointly learning topics in sentence embedding for document summarization}.
\newblock \emph{{IEEE} Trans. Knowl. Data Eng.}, 32(4):688--699.

\bibitem[{Geroda et~al.(2023)Geroda, Pane et~al.}]{geroda2023analysis}
Godefridus~Bali Geroda, Widi~Syahtia Pane, et~al. 2023.
\newblock An analysis language style based on the level of formality according to martin joos theory.
\newblock \emph{Inquest Journal}, 1(02):163--174.

\bibitem[{Halliday(1989)}]{Halliday1989SpokenAW}
Michael~A.K. Halliday. 1989.
\newblock \href {https://api.semanticscholar.org/CorpusID:58810022} {Spoken and written language}.

\bibitem[{Hu et~al.(2022)Hu, Lee, Aggarwal, and Zhang}]{hu2022text}
Zhiqiang Hu, Roy Ka-Wei Lee, Charu~C Aggarwal, and Aston Zhang. 2022.
\newblock Text style transfer: A review and experimental evaluation.
\newblock \emph{ACM SIGKDD Explorations Newsletter}, 24(1):14--45.

\bibitem[{Jin et~al.(2022)Jin, Jin, Hu, Vechtomova, and Mihalcea}]{i:1-TST-survey}
Di~Jin, Zhijing Jin, Zhiting Hu, Olga Vechtomova, and Rada Mihalcea. 2022.
\newblock \href {https://doi.org/10.1162/coli_a_00426} {Deep learning for text style transfer: A survey}.
\newblock \emph{Computational Linguistics}, 48(1):155--205.

\bibitem[{Kedrowicz and Taylor(2016)}]{Kedrowicz2016ShiftingRN}
April~A. Kedrowicz and Julie Taylor. 2016.
\newblock \href {https://api.semanticscholar.org/CorpusID:148248231} {Shifting rhetorical norms and electronic eloquence}.
\newblock \emph{Journal of Business and Technical Communication}, 30:352 -- 377.

\bibitem[{Krippendorff(2011)}]{Krippendorff2011ComputingKA}
Klaus Krippendorff. 2011.
\newblock \href {https://api.semanticscholar.org/CorpusID:59901023} {Computing krippendorff's alpha-reliability}.

\bibitem[{Lai et~al.(2021)Lai, Toral, and Nissim}]{Thank_you_BART}
Huiyuan Lai, Antonio Toral, and Malvina Nissim. 2021.
\newblock \href {https://doi.org/10.18653/v1/2021.acl-short.62} {Thank you {BART}! rewarding pre-trained models improves formality style transfer}.
\newblock In \emph{Proceedings of the 59th Annual Meeting of the Association for Computational Linguistics and the 11th International Joint Conference on Natural Language Processing (Volume 2: Short Papers)}, pages 484--494, Online. Association for Computational Linguistics.

\bibitem[{Lai et~al.(2023)Lai, Toral, and Nissim}]{lai2023multidimensional}
Huiyuan Lai, Antonio Toral, and Malvina Nissim. 2023.
\newblock Multidimensional evaluation for text style transfer using chatgpt.
\newblock \emph{arXiv preprint arXiv:2304.13462}.

\bibitem[{Li et~al.(2024{\natexlab{a}})Li, Yang, Bai, Zhou, Li, Sun, Liu, Si, Ye, Wu, Lin, Xu, Bowen, Feng, Gao, and Huang}]{li-etal-2024-fundamental}
Jiawei Li, Yizhe Yang, Yu~Bai, Xiaofeng Zhou, Yinghao Li, Huashan Sun, Yuhang Liu, Xingpeng Si, Yuhao Ye, Yixiao Wu, Yiguan Lin, Bin Xu, Ren Bowen, Chong Feng, Yang Gao, and Heyan Huang. 2024{\natexlab{a}}.
\newblock \href {https://doi.org/10.18653/v1/2024.acl-long.599} {Fundamental capabilities of large language models and their applications in domain scenarios: A survey}.
\newblock In \emph{Proceedings of the 62nd Annual Meeting of the Association for Computational Linguistics (Volume 1: Long Papers)}, pages 11116--11141, Bangkok, Thailand. Association for Computational Linguistics.

\bibitem[{Li et~al.(2024{\natexlab{b}})Li, Miao, Huang, and Gao}]{Word_Matters}
Yinghao Li, Siyu Miao, Heyan Huang, and Yang Gao. 2024{\natexlab{b}}.
\newblock \href {https://doi.org/10.18653/V1/2024.ACL-LONG.715} {Word matters: What influences domain adaptation in summarization?}
\newblock In \emph{Proceedings of the 62nd Annual Meeting of the Association for Computational Linguistics (Volume 1: Long Papers), {ACL} 2024, Bangkok, Thailand, August 11-16, 2024}, pages 13236--13249. Association for Computational Linguistics.

\bibitem[{Lin and Och(2004)}]{lin2004automatic}
Chin-Yew Lin and Franz~Josef Och. 2004.
\newblock Automatic evaluation of machine translation quality using longest common subsequence and skip-bigram statistics.
\newblock In \emph{Proceedings of the 42nd Annual Meeting of the Association for Computational Linguistics (ACL-04)}, pages 605--612.

\bibitem[{Madaan et~al.(2020)Madaan, Setlur, Parekh, P{\'o}czos, Neubig, Yang, Salakhutdinov, Black, and Prabhumoye}]{politeness}
Aman Madaan, Amrith~Rajagopal Setlur, Tanmay Parekh, Barnab{\'a}s P{\'o}czos, Graham Neubig, Yiming Yang, Ruslan Salakhutdinov, Alan~W. Black, and Shrimai Prabhumoye. 2020.
\newblock \href {https://api.semanticscholar.org/CorpusID:215811473} {Politeness transfer: A tag and generate approach}.
\newblock In \emph{Annual Meeting of the Association for Computational Linguistics}.

\bibitem[{McCroskey et~al.(2003)McCroskey, Wrench, and Richmond}]{mccroskey2003principles}
James~C. McCroskey, Jason~S. Wrench, and Virginia~Peck Richmond. 2003.
\newblock \emph{Principles of Public Speaking}.
\newblock The College Network, Indianapolis, IN.

\bibitem[{McDonald and Pustejovsky(1985)}]{mcdonald1985computational}
David~D McDonald and James Pustejovsky. 1985.
\newblock A computational theory of prose style for natural language generation.
\newblock In \emph{Second Conference of the European Chapter of the Association for Computational Linguistics}.

\bibitem[{Morande and Amini(2023)}]{Digital_Persona}
Swapnil Morande and Mitra Amini. 2023.
\newblock \href {https://doi.org/10.32388/0QI028} {Digital persona: Reflection on the power of generative ai for customer profiling in social media marketing}.

\bibitem[{OpenAI(2022)}]{chatGPT}
OpenAI. 2022.
\newblock \href {https:// openai.com/blog/chatgpt} {Introducing chatgpt}.

\bibitem[{Papineni et~al.(2002)Papineni, Roukos, Ward, and Zhu}]{papineni2002bleu}
Kishore Papineni, Salim Roukos, Todd Ward, and Wei-Jing Zhu. 2002.
\newblock Bleu: a method for automatic evaluation of machine translation.
\newblock In \emph{Proceedings of the 40th annual meeting of the Association for Computational Linguistics}, pages 311--318.

\bibitem[{Rao and Tetreault(2018)}]{rao-tetreault-2018-dear}
Sudha Rao and Joel Tetreault. 2018.
\newblock \href {https://doi.org/10.18653/v1/N18-1012} {Dear sir or madam, may {I} introduce the {GYAFC} dataset: Corpus, benchmarks and metrics for formality style transfer}.
\newblock In \emph{Proceedings of the 2018 Conference of the North {A}merican Chapter of the Association for Computational Linguistics: Human Language Technologies, Volume 1 (Long Papers)}, pages 129--140, New Orleans, Louisiana. Association for Computational Linguistics.

\bibitem[{Ratner and Gleason(2004)}]{Psycholinguistics}
N.B. Ratner and J.B. Gleason. 2004.
\newblock \href {https://doi.org/https://doi.org/10.1016/B978-008045046-9.01893-3} {Psycholinguistics}.
\newblock In Larry~R. Squire, editor, \emph{Encyclopedia of Neuroscience}, pages 1199--1204. Academic Press, Oxford.

\bibitem[{Sellam et~al.(2020{\natexlab{a}})Sellam, Das, and Parikh}]{sellam-etal-2020-bleurt}
Thibault Sellam, Dipanjan Das, and Ankur Parikh. 2020{\natexlab{a}}.
\newblock \href {https://doi.org/10.18653/v1/2020.acl-main.704} {{BLEURT}: Learning robust metrics for text generation}.
\newblock In \emph{Proceedings of the 58th Annual Meeting of the Association for Computational Linguistics}, pages 7881--7892, Online. Association for Computational Linguistics.

\bibitem[{Sellam et~al.(2020{\natexlab{b}})Sellam, Das, and Parikh}]{bleurt}
Thibault Sellam, Dipanjan Das, and Ankur~P. Parikh. 2020{\natexlab{b}}.
\newblock \href {https://arxiv.org/abs/2004.04696} {Bleurt: Learning robust metrics for text generation}.
\newblock In \emph{ACL}.

\bibitem[{Sheikha and Inkpen(2011)}]{sheikha2011generation}
Fadi~Abu Sheikha and Diana Inkpen. 2011.
\newblock Generation of formal and informal sentences.
\newblock In \emph{Proceedings of the 13th European Workshop on Natural Language Generation}, pages 187--193.

\bibitem[{Shen et~al.(2017{\natexlab{a}})Shen, Lei, Barzilay, and Jaakkola}]{i:2-sentiment}
Tianxiao Shen, Tao Lei, Regina Barzilay, and Tommi Jaakkola. 2017{\natexlab{a}}.
\newblock \href {https://proceedings.neurips.cc/paper_files/paper/2017/file/2d2c8394e31101a261abf1784302bf75-Paper.pdf} {Style transfer from non-parallel text by cross-alignment}.
\newblock In \emph{Advances in Neural Information Processing Systems}, volume~30. Curran Associates, Inc.

\bibitem[{Shen et~al.(2017{\natexlab{b}})Shen, Lei, Barzilay, and Jaakkola}]{shen2017style}
Tianxiao Shen, Tao Lei, Regina Barzilay, and Tommi Jaakkola. 2017{\natexlab{b}}.
\newblock Style transfer from non-parallel text by cross-alignment.
\newblock \emph{Advances in neural information processing systems}, 30.

\bibitem[{Touvron et~al.(2023)Touvron, Martin, Stone, Albert, Almahairi, Babaei, Bashlykov, Batra, Bhargava, Bhosale, Bikel, Blecher, Ferrer, Chen, Cucurull, Esiobu, Fernandes, Fu, Fu, Fuller, Gao, Goswami, Goyal, Hartshorn, Hosseini, Hou, Inan, Kardas, Kerkez, Khabsa, Kloumann, Korenev, Koura, Lachaux, Lavril, Lee, Liskovich, Lu, Mao, Martinet, Mihaylov, Mishra, Molybog, Nie, Poulton, Reizenstein, Rungta, Saladi, Schelten, Silva, Smith, Subramanian, Tan, Tang, Taylor, Williams, Kuan, Xu, Yan, Zarov, Zhang, Fan, Kambadur, Narang, Rodriguez, Stojnic, Edunov, and Scialom}]{i:1-LLaMA-2}
Hugo Touvron, Louis Martin, Kevin Stone, Peter Albert, Amjad Almahairi, Yasmine Babaei, Nikolay Bashlykov, Soumya Batra, Prajjwal Bhargava, Shruti Bhosale, Dan Bikel, Lukas Blecher, Cristian~Canton Ferrer, Moya Chen, Guillem Cucurull, David Esiobu, Jude Fernandes, Jeremy Fu, Wenyin Fu, Brian Fuller, Cynthia Gao, Vedanuj Goswami, Naman Goyal, Anthony Hartshorn, Saghar Hosseini, Rui Hou, Hakan Inan, Marcin Kardas, Viktor Kerkez, Madian Khabsa, Isabel Kloumann, Artem Korenev, Punit~Singh Koura, Marie-Anne Lachaux, Thibaut Lavril, Jenya Lee, Diana Liskovich, Yinghai Lu, Yuning Mao, Xavier Martinet, Todor Mihaylov, Pushkar Mishra, Igor Molybog, Yixin Nie, Andrew Poulton, Jeremy Reizenstein, Rashi Rungta, Kalyan Saladi, Alan Schelten, Ruan Silva, Eric~Michael Smith, Ranjan Subramanian, Xiaoqing~Ellen Tan, Binh Tang, Ross Taylor, Adina Williams, Jian~Xiang Kuan, Puxin Xu, Zheng Yan, Iliyan Zarov, Yuchen Zhang, Angela Fan, Melanie Kambadur, Sharan Narang, Aurelien Rodriguez, Robert Stojnic, Sergey Edunov, and Thomas
  Scialom. 2023.
\newblock \href {http://arxiv.org/abs/2307.09288} {Llama 2: Open foundation and fine-tuned chat models}.

\bibitem[{Wang et~al.(2023)Wang, Peng, Que, Liu, Zhou, Wu, Guo, Gan, Ni, Zhang et~al.}]{wang2023rolellm}
Zekun~Moore Wang, Zhongyuan Peng, Haoran Que, Jiaheng Liu, Wangchunshu Zhou, Yuhan Wu, Hongcheng Guo, Ruitong Gan, Zehao Ni, Man Zhang, et~al. 2023.
\newblock Rolellm: Benchmarking, eliciting, and enhancing role-playing abilities of large language models.
\newblock \emph{arXiv preprint arXiv:2310.00746}.

\bibitem[{Yang et~al.(2023)Yang, Sun, Li, Liu, Li, Liu, Huang, and Gao}]{yang2023mindllm}
Yizhe Yang, Huashan Sun, Jiawei Li, Runheng Liu, Yinghao Li, Yuhang Liu, Heyan Huang, and Yang Gao. 2023.
\newblock \href {http://arxiv.org/abs/2310.15777} {Mindllm: Pre-training lightweight large language model from scratch, evaluations and domain applications}.

\bibitem[{Yang et~al.(2018)Yang, Hu, Dyer, Xing, and Berg-Kirkpatrick}]{yang2018unsupervised}
Zichao Yang, Zhiting Hu, Chris Dyer, Eric~P Xing, and Taylor Berg-Kirkpatrick. 2018.
\newblock Unsupervised text style transfer using language models as discriminators.
\newblock \emph{Advances in Neural Information Processing Systems}, 31.

\bibitem[{Zhan et~al.(2022)Zhan, Gao, Bai, and Liu}]{two-stage}
Jiaao Zhan, Yang Gao, Yu~Bai, and Qianhui Liu. 2022.
\newblock \href {https://doi.org/10.24963/ijcai.2022/623} {Stage-wise stylistic headline generation: Style generation and summarized content insertion}.
\newblock In \emph{Proceedings of the Thirty-First International Joint Conference on Artificial Intelligence, {IJCAI-22}}, pages 4489--4495. International Joint Conferences on Artificial Intelligence Organization.
\newblock Main Track.

\bibitem[{Zhang et~al.(2024)Zhang, Zeng, Wang, and Lu}]{Zhang2024TinyLlamaAO}
Peiyuan Zhang, Guangtao Zeng, Tianduo Wang, and Wei Lu. 2024.
\newblock \href {https://api.semanticscholar.org/CorpusID:266755802} {Tinyllama: An open-source small language model}.
\newblock \emph{ArXiv}, abs/2401.02385.

\bibitem[{Zhang et~al.(2019)Zhang, Kishore, Wu, Weinberger, and Artzi}]{zhang2019bertscore}
Tianyi Zhang, Varsha Kishore, Felix Wu, Kilian~Q Weinberger, and Yoav Artzi. 2019.
\newblock Bertscore: Evaluating text generation with bert.
\newblock \emph{arXiv preprint arXiv:1904.09675}.

\end{thebibliography}
\onecolumn
\appendix
\section{PSST Dataset}
\label{Dataset}
\subsection{Target Dataset: Public Speaking Style Data from Real Scenarios}
\label{Target_Dataset}
\paragraph{Ted Talks} Ted Talks is a series of short, powerful presentations that share ideas and insights on creativity, science, and culture. The raw dataset can be found here\footnote{https://huggingface.co/datasets/iwslt2017/viewer/iwslt2017-
en-zh}. Processed texts can be found in our github. We use 20 samples for prior fine-grained analysis in Section~\ref{Features_Analysis}
\paragraph{Political Speeches\footnote{https://www.americanrhetoric.com/}} Political speeches are crafted to persuade, using emotional appeals and persuasive language to emphasize points, highlight policies, and influence public opinion. We use 20 samples for prior fine-grained analysis in Section~\ref{Features_Analysis}
\paragraph{Academic Presentations\footnote{https://iwslt.org/2023/multilingual}} Academic presentations introduce scientific research to the audience, focusing on engaging the audience and making complex content easier to understand. We use 20 samples for prior fine-grained analysis in Section~\ref{Features_Analysis}
\paragraph{Lecture Transcripts\footnote{https://www.webpages.uidaho.edu/psyc390/index.htm}} Lecture transcripts are delivered by teachers in class, designed to facilitate student understanding and engagement over an extended period. We use 20 samples for prior fine-grained analysis in Section~\ref{Features_Analysis}

It is important to note that we utilize the above data to analyze general stylistic characteristics. Furthermore, in this paper, we primarily employ TED data as a representative example to demonstrate the effectiveness of our evaluation pipeline.  Likewise, the same evaluation pipeline can be applied to data from other scenarios to achieve comparable results.

\subsection{Source Dataset: Official-Style Data Collected of PSST}
\label{Source_Dataset}
The sources of data utilized for official-style texts are as follows:
\paragraph{News} The data from the news category is sourced from the Fake and Real News dataset available on Kaggle\footnote{https://www.kaggle.com/datasets/clmentbisaillon/fake-and-real-news-dataset}. This dataset comprises entire news articles from Reuters and was intended for news classification, including both real and fake news. We selected the subset of the real news.  

\paragraph{Paper Abstracts} The dataset from the research paper abstract category is sourced from the arXiv Dataset on Kaggle\footnote{https://www.kaggle.com/datasets/Cornell-University/arxiv}. This dataset is a repository containing a substantial number of articles from arXiv, including details like article titles, authors, categories, abstracts, and full-text PDFs. For our purposes, we have extracted the abstract portions. 

\paragraph{Wikipedia} The dataset from the encyclopedia category is obtained from Hugging Face's wikitext dataset\footnote{https://huggingface.co/datasets/wikitext}, which is a collection of over 100 million tokens extracted from the set of verified Good and Featured articles on Wikipedia.

\subsection{Data Processing and Final PSST Dataset}
\begin{table}[h]
    \centering
    \begin{tabular}{c|c|cccc}
    \toprule
         \multirow{2}{*}{\textbf{Token Nums}}&\textbf{Targe Dataset}&\multicolumn{4}{c}{\textbf{Source Dataset}}\\
         \cline{2-6}
    &Ted Talks&News&Abstract&Wiki&Total\\
    \midrule
    $400\pm100$&32&40&40&40&120\\
    $800\pm200$&99&40&40&40&120\\
    $1200\pm200$&144&60&-&60&120\\
    \bottomrule
    \end{tabular}
    \caption{Statistics of the final source and target datasets used for PSST.}
    \label{tab:final_dataset_statistics}
\end{table}
To ensure the reliability of our evaluation and comprehensively assess the model's stylization capabilities, we further select the source and target datasets by filtering based on token counts, ensuring comparable lengths between the two. Given the severe semantic losses observed in current LLMs, we advocate for studying their complex style modeling capabilities within bounds (we set a maximum test token number limit of 1400) that are reasonable given their current performance levels. The final dataset we used for PSST is shown in Table~\ref{tab:final_dataset_statistics}.

\section{Datasets and Training Details of Style Evaluation Modeling}
\label{Dataset_Training_Details}
\subsection{Statistics of Sentence-Level Style-Strength Scoring Training Data}
To distill the sentence-level style strength evaluation capability of LLMs into small-scale models, we ask \text{gpt-3.5-turbo} to generate training data shown in Table~\ref{tab:evaluation_mdoeling_dataset}. For detailed information on prompts, please refer to Appendix~\ref{sec:prompt_for_generate_data}.
\begin{table}[ht]
    \centering
    \begin{tabular}{ccc}
    \toprule
         \textbf{Dimension}&\textbf{Train set}&\textbf{validation set}  \\
    \midrule
         interactivity&7769 &864 \\
         emotionality&7884 &876 \\
         vividness&7942 &883 \\
         orality&7839 &871 \\
    \bottomrule
    \end{tabular}
    \caption{Statistics of sentence-level style-strength scoring training data}
    \label{tab:evaluation_mdoeling_dataset}
\end{table}
\subsection{Training Details of Sentence-Level Scoring Models}
For each evaluation dimension, we fine-tuned TinyLlama-1.1B~\citep{Zhang2024TinyLlamaAO} as the evaluation model. The training parameters are detailed in Table~\ref{tab:training_details}.
\begin{table}[hbpt]
    \centering
    \begin{tabular}{ccccc}
    \toprule
        \textbf{Optimizer}&\textbf{LR\_Scheduler}&\textbf{Learning Rate}&\textbf{Batch Size}&\textbf{Epochs}  \\
    \midrule
    AdamW& \makecell[c]{Warmup=0.03\\Decay="cosine"}&$2\times 10^{-5}$&256&6\\ 
    \bottomrule
    \end{tabular}
    \caption{Training Details of evaluation modeling.}
    \label{tab:training_details}
\end{table}

\section{Experiments}
\label{sec:additional_exps}
\subsection{Results of Different Text Lengths in PSST}
\begin{figure*}[h]
    \centering
    \includegraphics[scale=0.3]{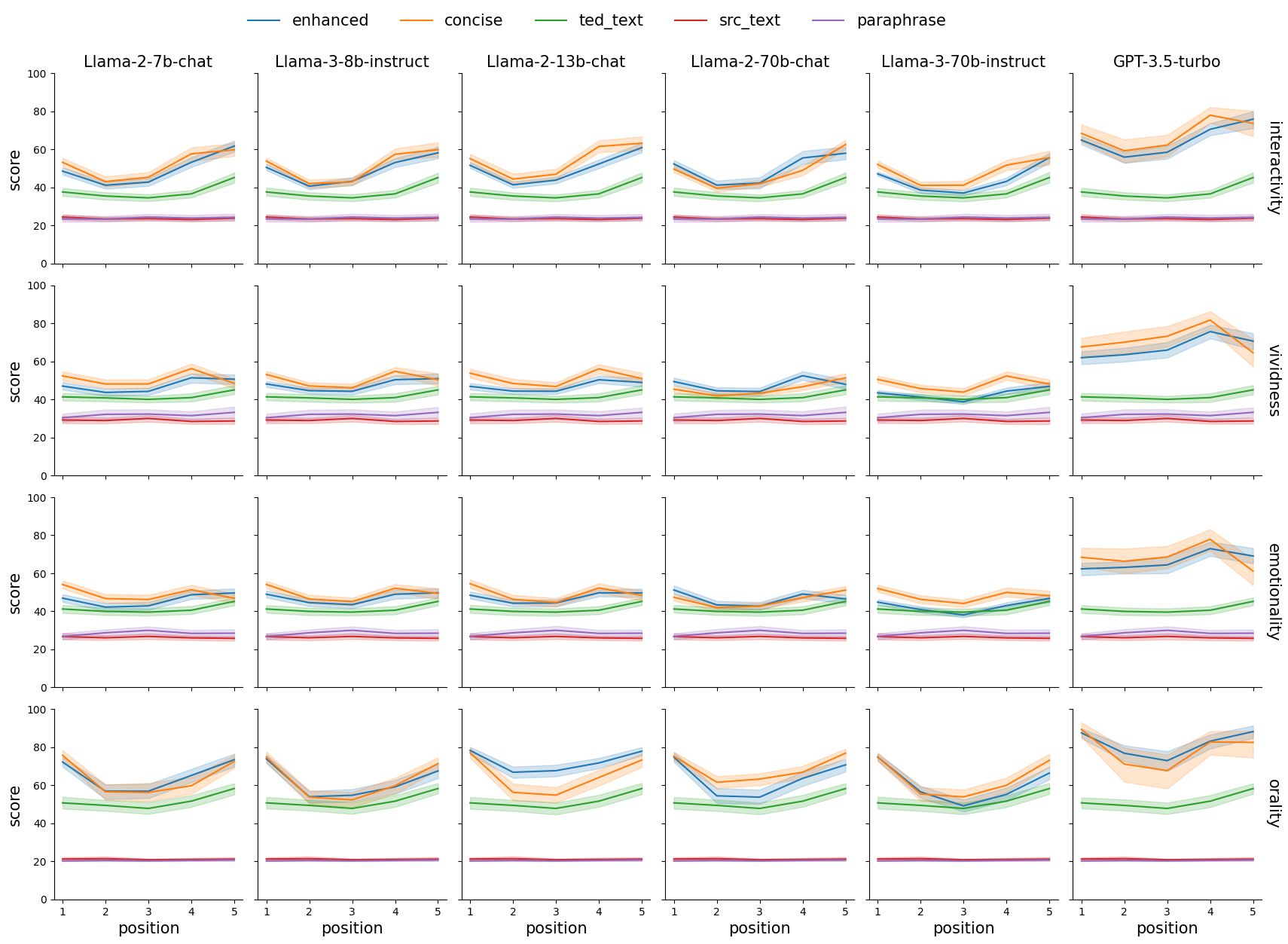}
    \caption{Style Strength Distribution of Passages Transferred by Different LLMs($800\pm200$ tokens).}
    \label{fig:style_strength_distribution_800}
\end{figure*}

\begin{figure*}[h]
    \centering
    \includegraphics[scale=0.3]{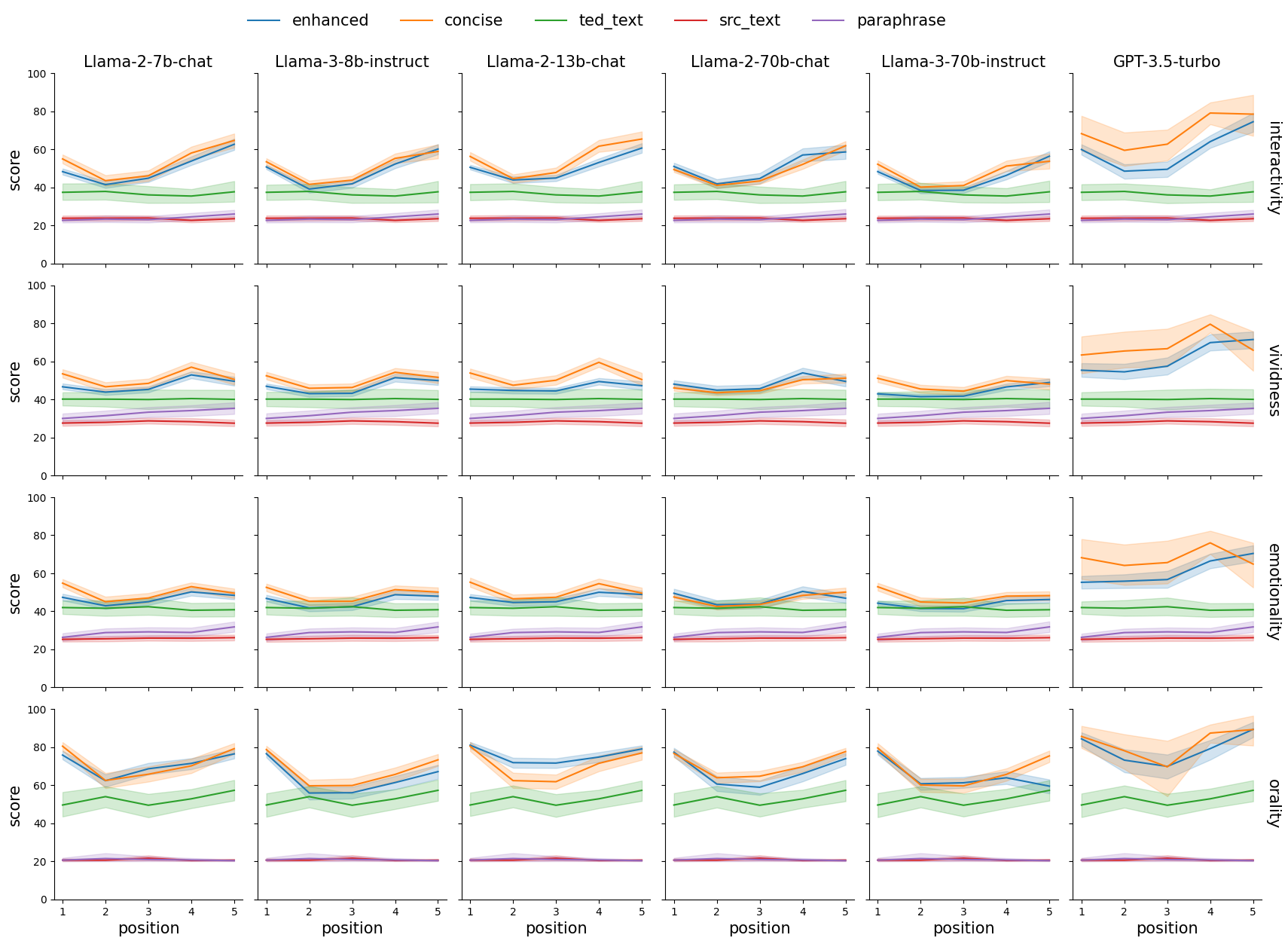}
    \caption{Style Strength Distribution of Passages Transferred by Different LLMs($400\pm100$ tokens).}
    \label{fig:style_strength_distribution_400}
\end{figure*}

\begin{figure*}[h]
    \centering
    \includegraphics[scale=0.3]{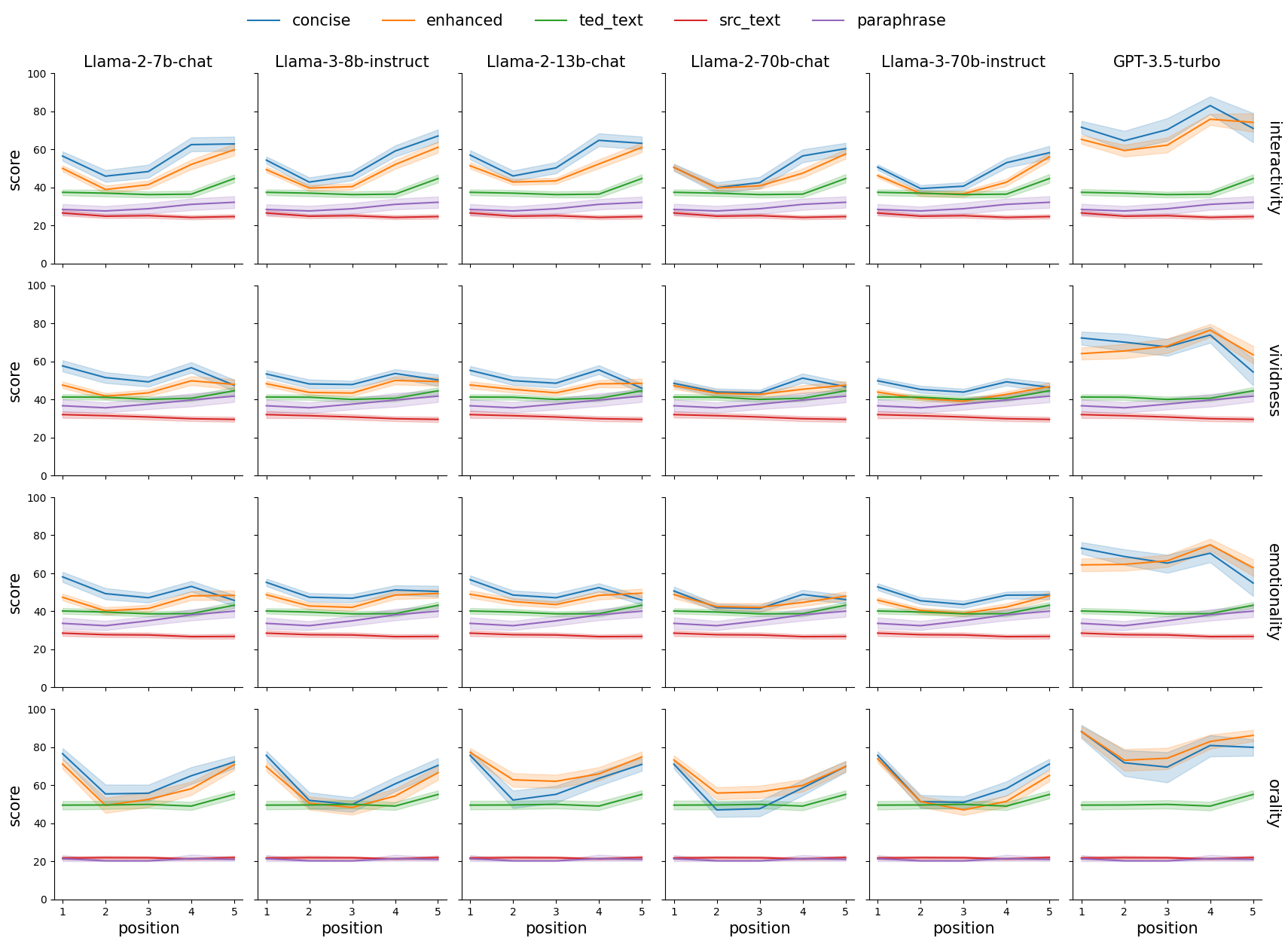}
    \caption{Style Strength Distribution of Passages Transferred by Different LLMs($1200\pm200$ tokens).}
    \label{fig:style_strength_distribution_1200}
\end{figure*}

In our investigation, we employed Public-Speaking Style Transfer (PSST) on text samples of three distinct lengths: $400\pm100$ tokens, $800\pm200$ tokens, and $1200\pm200$ tokens. The experiments aimed to evaluate the impact of text length on the efficacy of current LLMs. The results of these experiments are systematically presented in Table~\ref{tab:style_eval_res_800} \& Figure~\ref{fig:style_strength_distribution_800} (for $800\pm200$-token texts), Table~\ref{tab:style_eval_res_400} \& Figure~\ref{fig:style_strength_distribution_400} (for $400\pm100$-token texts) and Table~\ref{tab:style_eval_res_1200} \& Figure~\ref{fig:style_strength_distribution_1200}  (for $400\pm100$-token texts), respectively. These tables detail the performance metrics obtained, providing a clear comparative analysis of the outcomes across different text lengths.

\begin{table*}[h]\small
\centering
\begin{tabular}{lcccccccc}
\toprule
\multirow{2}{*}{\textbf{Model}}& \multicolumn{5}{c}{\textbf{Style Strength Score(\%)}}&\multicolumn{3}{c}{\textbf{Semantic Preservation(\%)}} \\
\cline{2-5}\cline{7-9}
& Interactivity & Vividness & Emotionality & Orality &  &Details&Logic&Average\\
\midrule
\rowcolor{mygray} \multicolumn{9}{c}{\textbf{\textit{Real Scenes}}}\\
Ted Talks&41.11 &42.67 &42.36 &51.50 & &-&-&- \\
Political Speech&47.56 &54.91 &53.57 &45.92 & &-&-&- \\
Course Teaching&29.72 &33.77 &29.70 &33.60 & &-&-&- \\
\midrule
\rowcolor{mygray} \multicolumn{9}{c}{\textbf{\textit{Baseline}}}\\
src\_text&23.54 &28.79 &26.20 &21.07&&97.00&94.33&95.66\\
paraphrase&27.22 &34.92 &31.42 &20.88&&88.67 &87.67 &88.17\\
\midrule
\rowcolor{mygray} \multicolumn{9}{c}{\textbf{\textit{Concise Prompt}}}\\
LLaMA-2-7b&51.60 &50.97 &49.66 &63.90&&69.00&69.00&69.00\\
LLaMA-3-8b&51.18 &51.00 &50.43 &63.28&&65.67&68.67&67.17\\
LLaMA-2-13b&54.36 &52.57&50.92&65.45& &66.33&62.67&64.50\\
LLaMA-2-70b &49.52 &47.93 &46.85 &63.11 &&71.67&67.00&69.34\\
LLaMA-3-70b &47.98&48.08 &48.29 &63.38 &&78.33&76.67&77.50\\
GPT-3.5&69.24 &71.39&68.79 &79.66 &&64.33&66.67&65.50\\
\midrule
\rowcolor{mygray} \multicolumn{9}{c}{\textbf{\textit{Enchanced Prompt}}}\\
LLaMA-2-7b&48.63&46.92&45.65&64.39&&74.00&69.67&71.84\\
LLaMA-3-8b&48.54&47.27&46.96&61.80&&71.00&68.00&69.50\\
LLaMA-2-13b&49.42&46.72&47.15&72.37&&70.00&72.33&71.16\\
LLaMA-2-70b&47.78&45.39&45.85&68.20&&72.00&70.67&71.34\\
LLaMA-3-70b&43.80&42.69&42.58&60.68&&84.33&80.67&82.50\\
GPT-3.5&64.27&66.89&65.46&79.58&&73.33&70.33&71.83\\
\bottomrule
\end{tabular}
\caption{Style Strength Evaluation Results($\textbf{800}\pm200$ tokens). \textbf{a.} The original score ranges from 0 to 5. \textbf{b.} \textbf{src\_text} means \textit{Source Text} and \textbf{paraphrase} means \textit{Paraphrased Text}, both are styleless. \textbf{c.} Scored by TinyLlama-1.1b(Reward).}
\label{tab:style_eval_res_800}
\end{table*}

\begin{table*}[h]\small
\centering
\begin{tabular}{lcccccccc}
\toprule
\multirow{2}{*}{\textbf{Model}}& \multicolumn{5}{c}{\textbf{Style Strength Score(\%)}}&\multicolumn{3}{c}{\textbf{Semantic Preservation(\%)}} \\
\cline{2-5}\cline{7-9}
& Interactivity & Vividness & Emotionality & Orality &  &QA1&QA2&Average\\
\midrule
\rowcolor{mygray} \multicolumn{9}{c}{\textbf{\textit{Baseline}}}\\
src\_text&23.38 &28.03 &25.82 &20.75& &92.67 &98.33 &95.50\\
paraphrase&27.56 &36.42 &31.87 &21.15& &89.67 &93.33 &91.50\\
TED&41.08 &40.11 &41.50 &52.50& &-&-&-\\
\midrule
\rowcolor{mygray} \multicolumn{9}{c}{\textbf{\textit{Concise Prompt}}}\\
LLaMA-2-7b &53.53 &51.57 &50.33 &70.10& &80.00 &77.33 &78.67 \\
LLaMA-2-13b &55.54 &53.12&51.80&70.80& &79.67 &81.00 &80.33 \\
LLaMA-2-70b &50.29 &48.83 &47.04 &66.48 & &81.67 &82.67 &82.17 \\
LLaMA-3-8b &50.96 &50.25 &48.89 &67.11& &81.67 &83.00 &82.33 \\
LLaMA-3-70b &48.71 &48.98 &48.82 &68.41 & &86.33 &85.33 &85.83 \\
GPT-3.5 &66.63&68.09&64.54 &77.74 & &80.00 &79.67 &79.83\\
\midrule
\rowcolor{mygray} \multicolumn{9}{c}{\textbf{\textit{Enchanced Prompt}}}\\
LLaMA-2-7b &49.69 &47.58 &47.00 &70.82 & &79.33 &82.67 &81.00 \\
LLaMA-2-13b &50.06 &46.08 &47.05 &75.67 & &79.00 &81.00 &80.00 \\
LLaMA-2-70b &48.89 &46.66 &46.11 &70.13 & &80.33 &83.00 &81.67 \\
LLaMA-3-8b &48.22 &46.63 &45.40 &63.51 & &83.00 &84.33 &83.67 \\
LLaMA-3-70b &44.98 &43.99 &43.70 &65.08 & &92.33 &92.33 &92.33 \\
GPT-3.5 &60.28 &62.90 &60.74 &76.42 & &82.00 &87.00 &84.50\\
\bottomrule
\end{tabular}
\caption{Style Strength Evaluation Results($\textbf{400}\pm100$ tokens). \textbf{a.} The original score ranges from 0 to 5. \textbf{b.} \textbf{src\_text} means \textit{Source Text} and \textbf{paraphrase} means \textit{Paraphrased Text}, both are styleless. \textbf{c.} Scored by TinyLlama-1.1b(Reward).}
\label{tab:style_eval_res_400}
\end{table*}

\begin{table}[h]
    \centering
    \begin{tabular}{lccc}
    \toprule
         \textbf{Metric}&\textbf{Kendall’s $\tau$} &\textbf{Spearman's $\rho$ }&\textbf{Krippendorff's $\alpha$}\\
    \midrule
         BLUERT&0.5111&0.6000&0.6067 \\
         BertScore-f1&0.5778&0.6667&0.6722 \\
         QA-based(ours)&\textbf{0.7122}&\textbf{0.7500}&\textbf{0.7542} \\
    \bottomrule
    \end{tabular}
    \caption{Correlation between different semantic preservation evaluation methods and human evaluation. (inter-annotator consistency: Krippendorff’s $\alpha$ = 0.7693)}
    \label{tab:Results_of_Correlation_Semantic_preservation}
\end{table}

\begin{table*}[h]\small
\centering
\begin{tabular}{lcccccccc}
\toprule
\multirow{2}{*}{\textbf{Model}}& \multicolumn{5}{c}{\textbf{Style Strength Score(\%)}}&\multicolumn{3}{c}{\textbf{Semantic Preservation(\%)}} \\
\cline{2-5}\cline{7-9}
& Interactivity & Vividness & Emotionality & Orality &  &QA1&QA2&Average\\
\midrule
\rowcolor{mygray} \multicolumn{9}{c}{\textbf{\textit{Baseline}}}\\
src\_text&25.12 &30.73 &27.48 &21.75& &98.00 &97.00 &97.50\\
paraphrase&31.25 &40.17 &36.20 &21.20 & &80.00 &84.33 &82.17\\
TED&38.28 &41.45 &40.07 &50.63 & &-&-&-\\
\midrule
\rowcolor{mygray} \multicolumn{9}{c}{\textbf{\textit{Concise Prompt}}}\\
LLaMA-2-7b &54.32 &52.47 &50.80 &63.68& &55.00 &61.00 &58.00\\
LLaMA-2-13b &56.67 &53.26&52.45&65.03& &55.33 &62.33 &58.83\\
LLaMA-2-70b &49.31 &46.69 &45.75 &58.16 & &66.67 &66.33 &66.50\\
LLaMA-3-8b &53.36 &51.18 &50.82 &61.52 & &64.67 &71.67 &68.17\\
LLaMA-3-70b &48.07 &47.32 &48.30 &61.76 & &74.33 &72.67 &73.50\\
GPT-3.5&71.13&68.12&66.01 &76.26 & &59.67 &59.33 &59.50 \\
\midrule
\rowcolor{mygray} \multicolumn{9}{c}{\textbf{\textit{Enchanced Prompt}}}\\
LLaMA-2-7b &47.86 &45.82 &44.83 &60.18& &67.67 &68.33 &68.00\\
LLaMA-2-13b &49.67 &46.52 &47.04 &68.50& &63.00 &68.67 &65.83\\
LLaMA-2-70b &46.82 &45.12 &45.20 &63.03 & &70.33 &71.33 &70.83\\
LLaMA-3-8b &47.86 &46.68 &46.16 &57.82 & &71.67 &73.67 &72.67\\
LLaMA-3-70b &43.16 &42.34 &43.00 &57.98 & &82.00 &82.33 &82.17\\
GPT-3.5 &66.38&68.49&67.06 &78.19 & &63.33 &67.67 &65.50\\
\bottomrule
\end{tabular}
\caption{Style Strength Evaluation Results($\textbf{1200}\pm100$ tokens). \textbf{a.} The original score ranges from 0 to 5. \textbf{b.} \textbf{src\_text} means \textit{Source Text} and \textbf{paraphrase} means \textit{Paraphrased Text}, both are styleless. \textbf{c.} Scored by TinyLlama-1.1b(Reward).}
\label{tab:style_eval_res_1200}
\end{table*}
\section{Prompts}
\label{sec:Prompts}
\subsection{Prompt for LLMs to Generate Sentence-Level Data to Train Evaluation Models}
\label{sec:prompt_for_generate_data}
\begin{figure}[ht]
    \centering
    \includegraphics[scale=0.45]{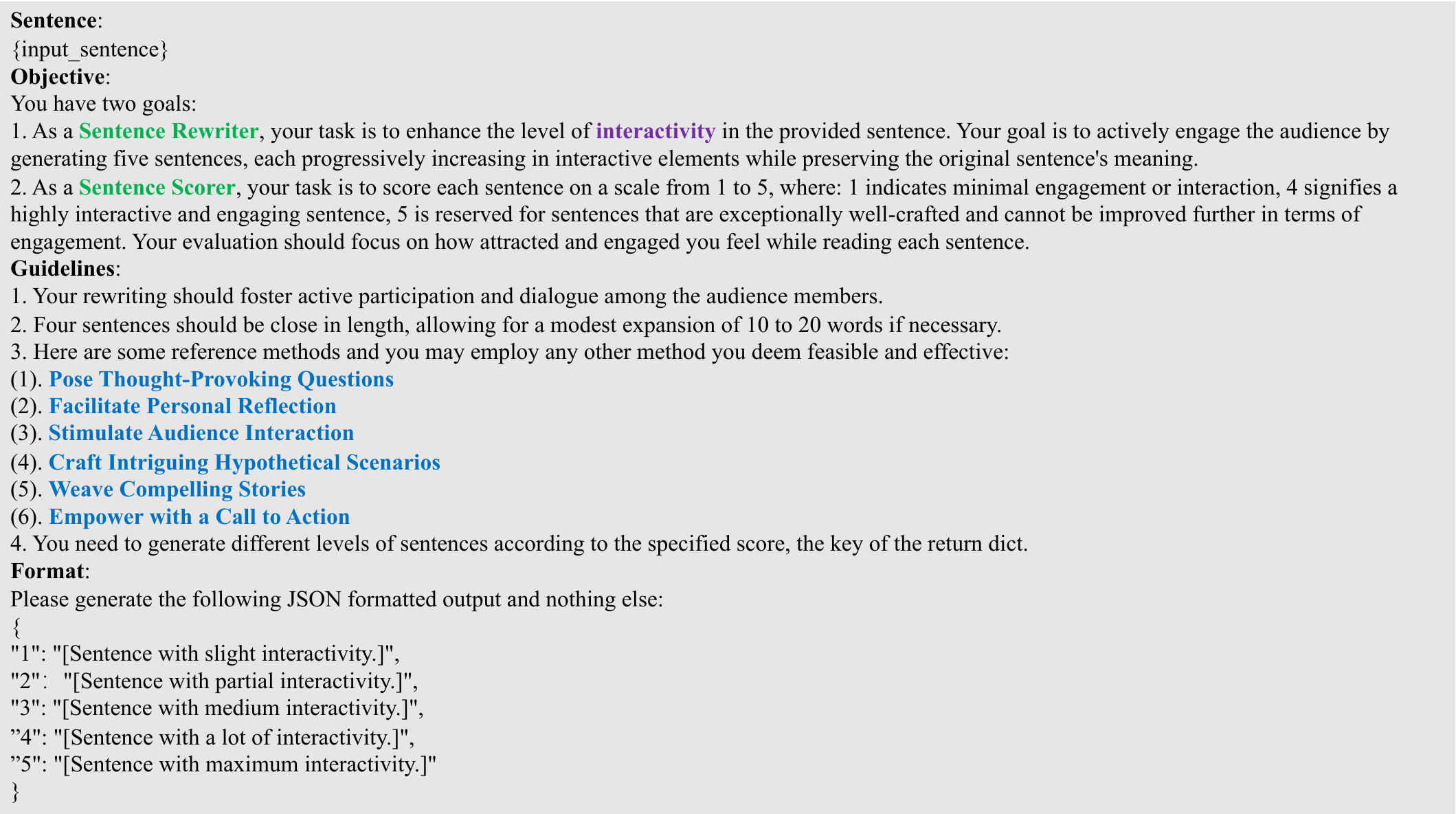}
    \caption{Prompt used for \texttt{gpt-3.5-turbo} to generate and score different sentence-level style strengths in interactivity dimension.}
    \label{fig:prompt4interactivity_data_gen}
\end{figure}

\begin{figure}[ht]
    \centering
    \includegraphics[scale=0.45]{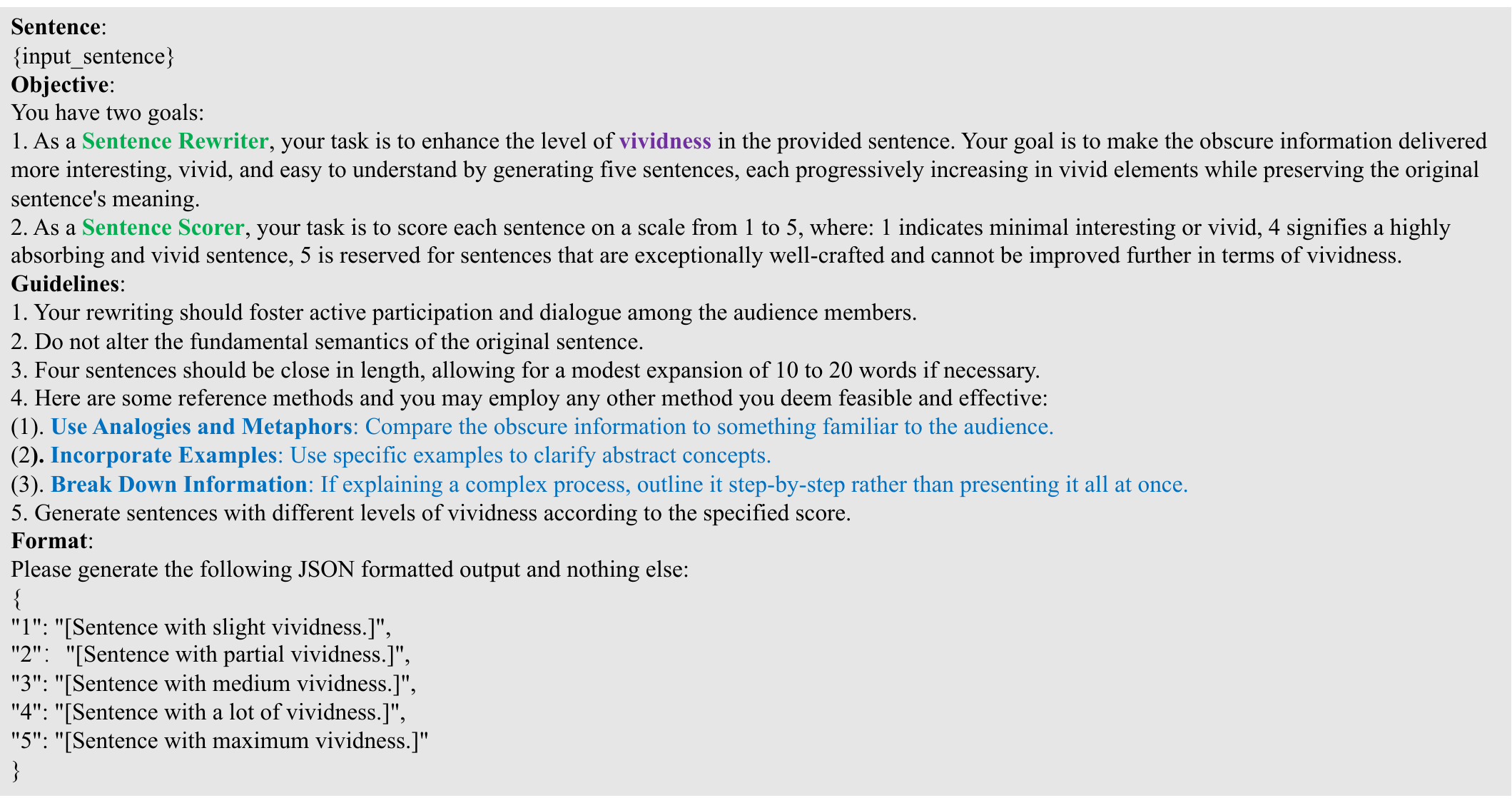}
    \caption{Prompt used for \texttt{gpt-3.5-turbo} to generate and score different sentence-level style strengths in the vividness dimension.}
    \label{fig:prompt4vividness_data_gen}
\end{figure}

\begin{figure}[ht]
    \centering
    \includegraphics[scale=0.45]{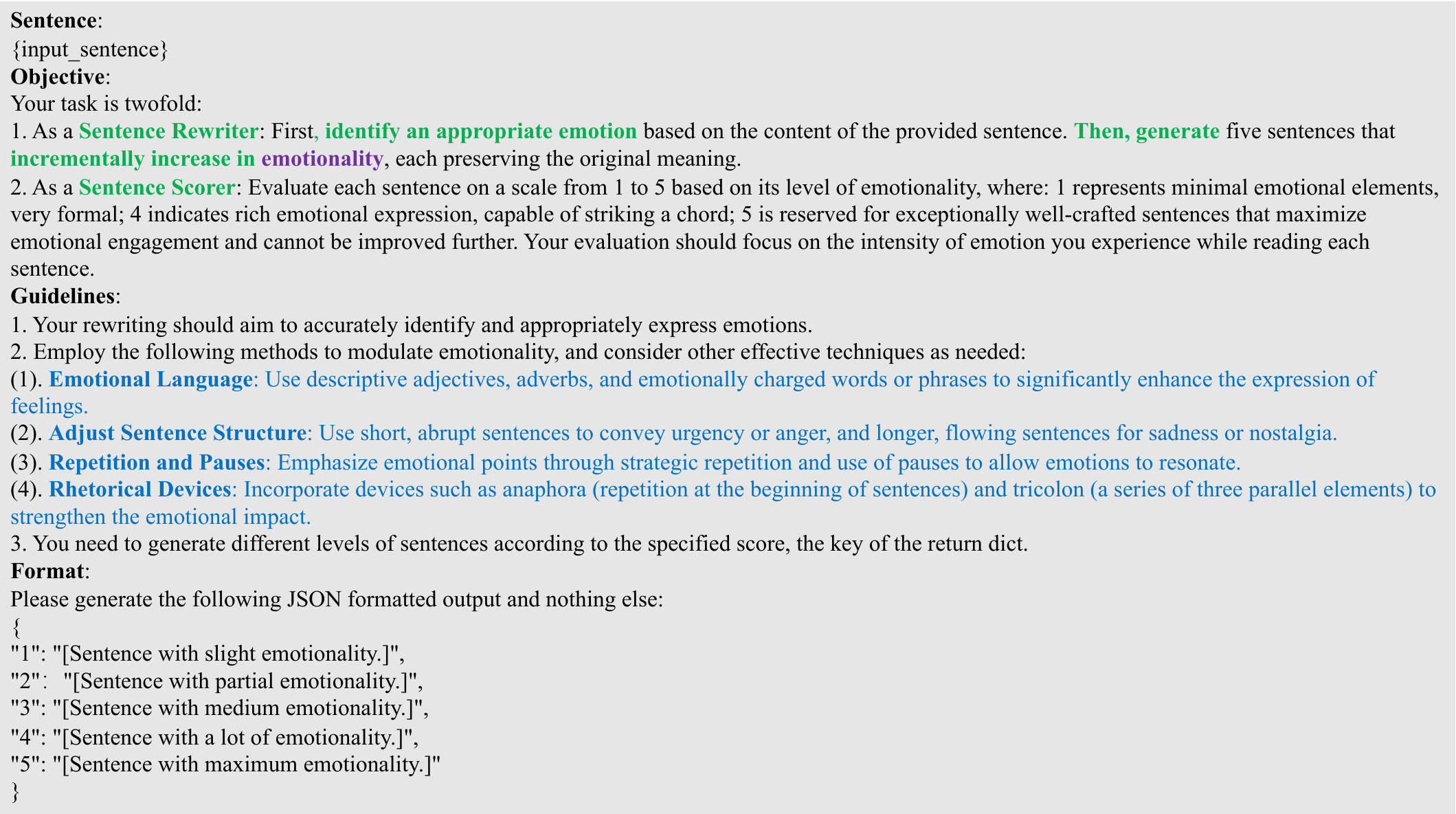}
    \caption{Prompt used for \texttt{gpt-3.5-turbo} to generate and score different sentence-level style strengths in the emotionality dimension. Particularly, we ask the \texttt{gpt-3.5-turbo} to identify the appropriate emotion before generating sentences and we opt not to restrict sentence length}
    \label{fig:prompt4emotionality_data_gen}
\end{figure}

\begin{figure}[ht]
    \centering
    \includegraphics[scale=0.45]{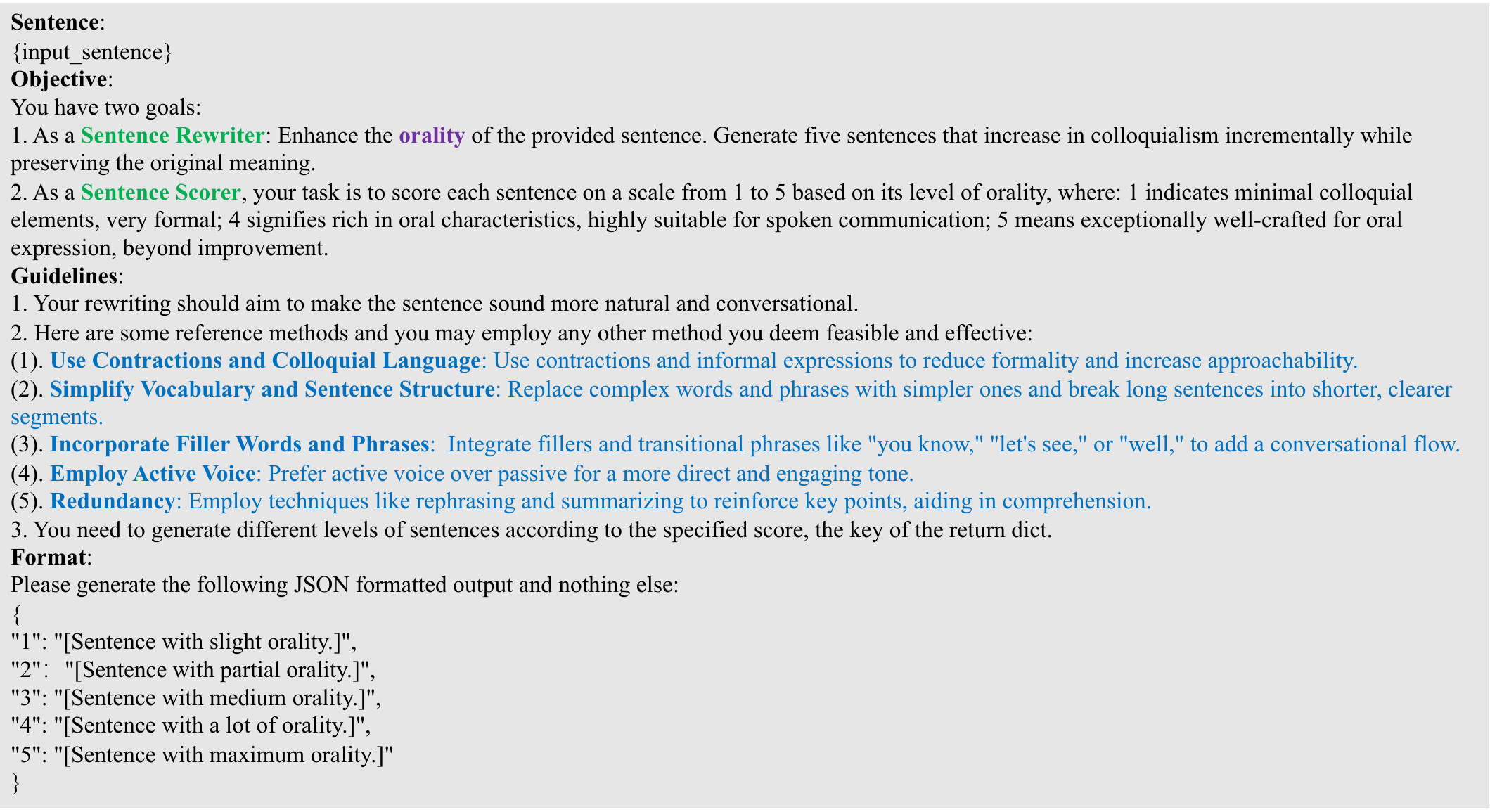}
    \caption{Prompt used for \texttt{gpt-3.5-turbo} to generate and score different sentence-level style strengths in orality dimension. We opt not to restrict sentence length due to that alterations in sentence patterns (e.g., simplifying complex structures) could significantly change sentence length. }
    \label{fig:prompt4orality_data_gen}
\end{figure}

We utilized the capabilities of GPT-3.5 to generate sentence-level data for training models to evaluate style strength. As depicted in Figures~\ref{fig:prompt4interactivity_data_gen}(interactivity), ~\ref{fig:prompt4emotionality_data_gen}(emotionality), \ref{fig:prompt4vividness_data_gen}(vividness), and \ref{fig:prompt4orality_data_gen}(orality), we ask the GPT-3.5 to both generate sentences with varying levels of stylistic intensity and concurrently assess the style strength of each sentence. We require the generated sentences to be consistent in meaning and close in length to minimize the impact of factors unrelated to style strength when scoring.

Particularly, for "emotionality", we ask the model to identify the appropriate emotion before generating sentences with different levels of emotionality. For "emotionality" and "orality," where the inclusion of additional emotional content and alterations in sentence patterns (e.g., simplifying complex structures) could significantly change sentence length, we opt not to restrict sentence length. 


\subsection{Prompts for LLMs to Generate QA-pairs for semantic preservation evaluation}
\label{sec:prompt_for_generate_QA_pairs}
\begin{figure}
    \centering
    \includegraphics[scale=0.45]{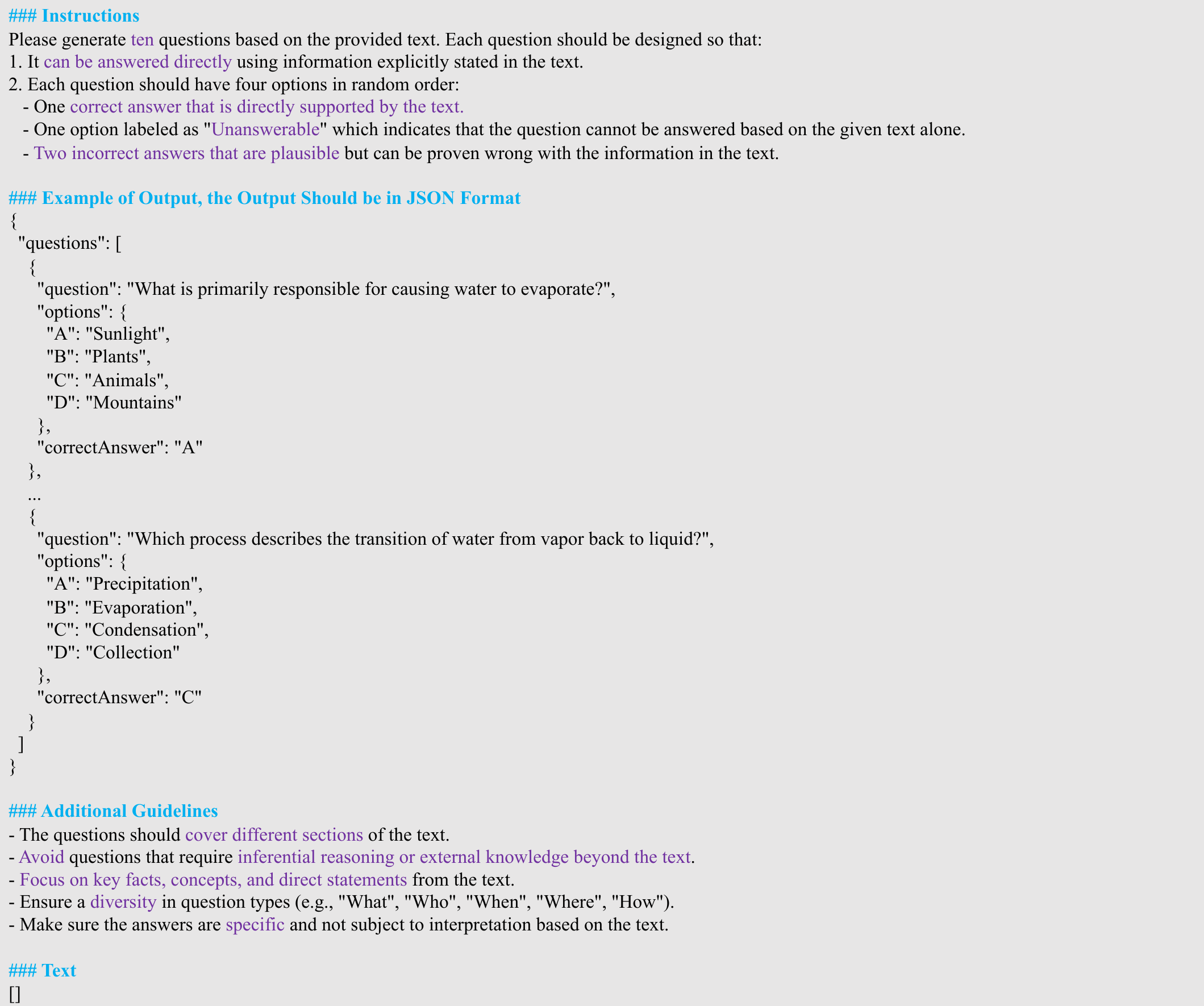}
    \caption{Prompts used to generate 10 question-answer pairs related to \textbf{key information}, aiming to evaluate the semantic preservation capabilities of LLMs. The \textcolor{purple}{purple} indicates specific guidance.}
    \label{fig:prompt4qa_pairs_generation_1}
\end{figure}

\begin{figure}
    \centering
    \includegraphics[scale=0.45]{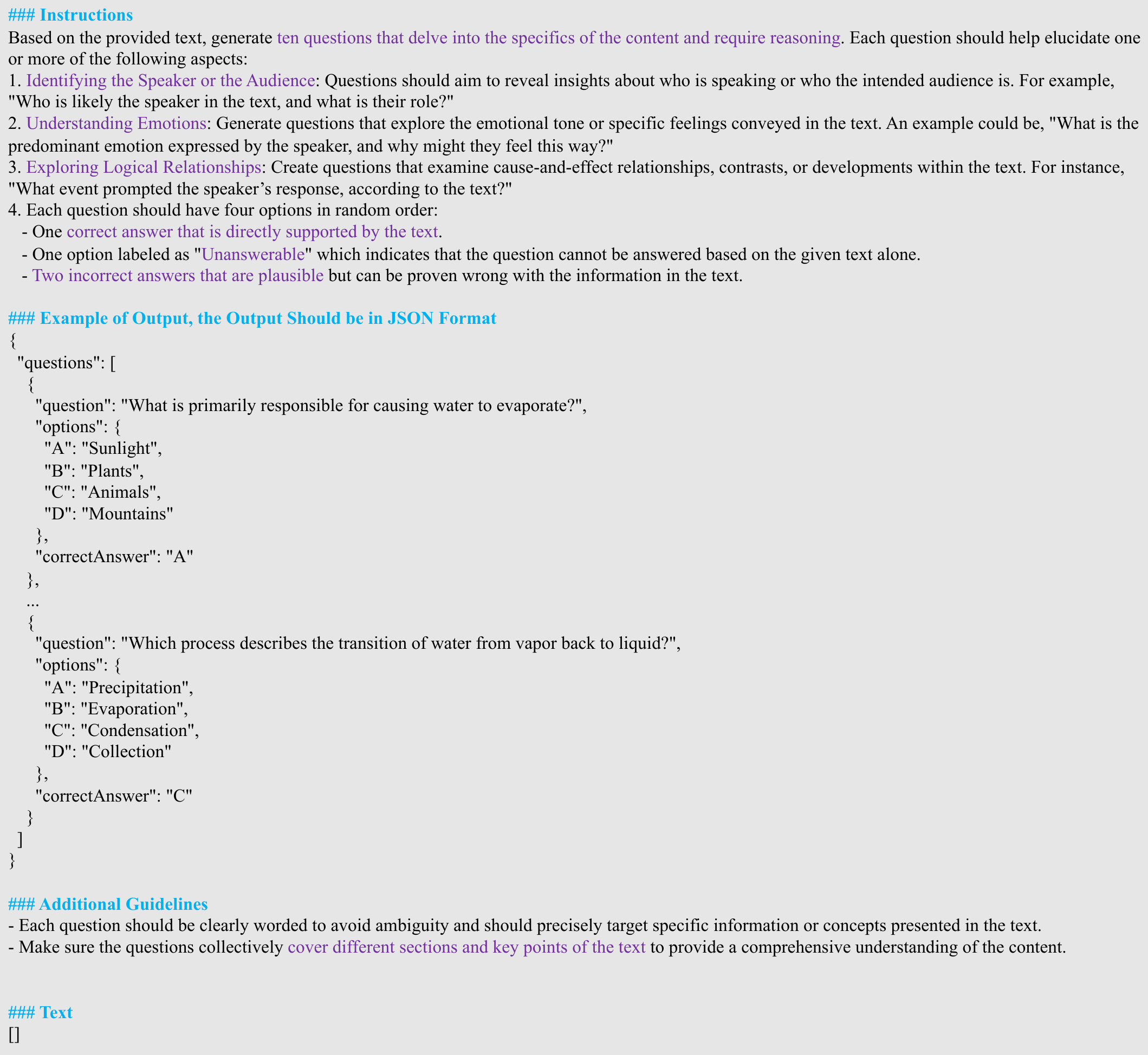}
    \caption{Prompt used to generate 10 question-answer pairs related to \textbf{logic and structure}, aiming to evaluate the semantic preservation capabilities of LLMs. The \textcolor{purple}{purple} indicates specific guidance.}
    \label{fig:prompt4qa_pairs_generation_2}
\end{figure}

Prompts used to generate QA pairs are shown in Figure~\ref{fig:prompt4qa_pairs_generation_1} and ~\ref{fig:prompt4qa_pairs_generation_2}.
\subsection{Prompt for LLMs to Perform PSST}
\label{sec:_prompt_for_PSST}
\begin{figure}
    \centering
    \includegraphics[scale=0.45]{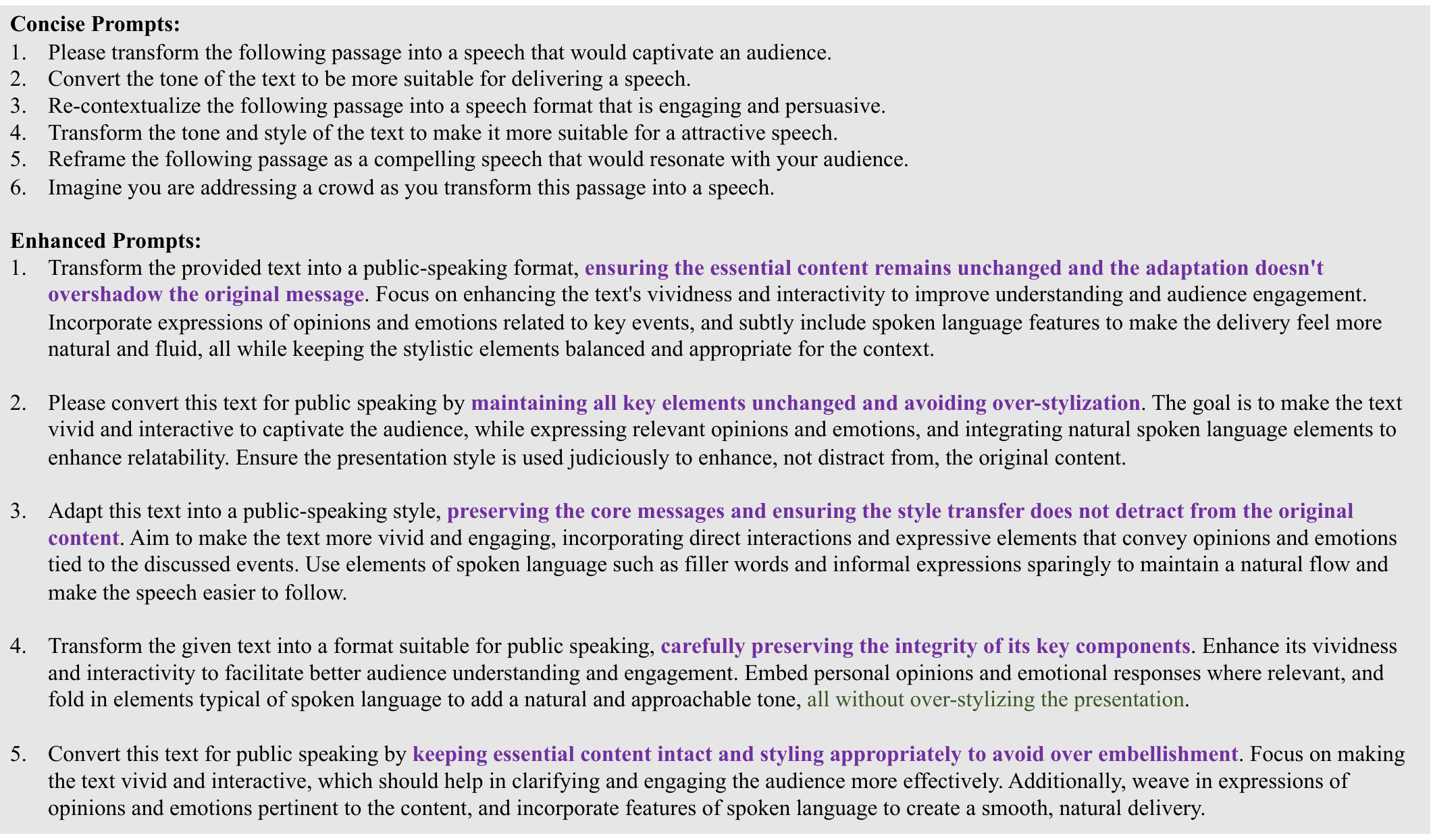}
    \caption{Concise and enhanced instructions to help LLMs perform PSST. The \textcolor{purple}{purple} indicates specific guidance.}
    \label{fig:prompt4LLMs2PSST}
\end{figure}

We design two prompt types for LLMs to conduct Text Speech-Style Transfer (PSST): a concise prompt and an enhanced prompt, illustrated in Figures~\ref{fig:prompt4LLMs2PSST}. The enhanced prompt incorporates emphasis, hints, and guidance pertaining to various speech-style characteristics. This is intended to assist LLMs in generating text with more appropriate and accurate stylistic features aligned with human preferences.

\subsection{Prompt for LLMs to Evaluate Style Strength}
\label{sec:Prompts_for_LLMs_evalaution_style}
\begin{figure}[ht]
    \centering
    \includegraphics[scale=0.45]{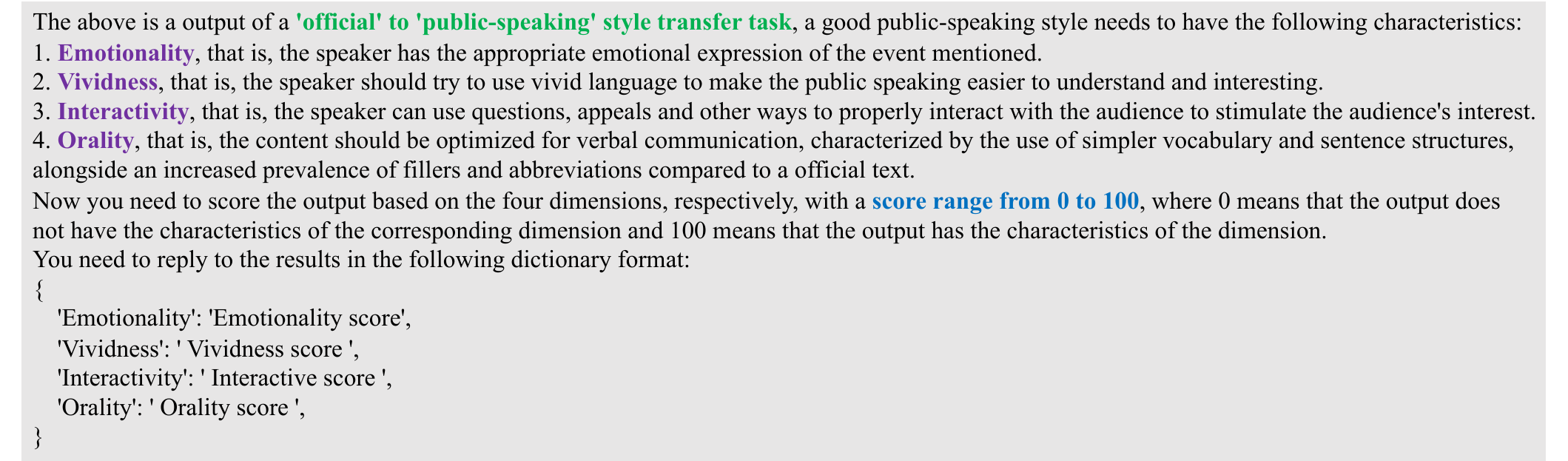}
    \caption{Prompt used for Llama 2-Chat-70B to score the public speaking style strength in four dimensions.}
    \label{fig:prompt_for_evaluation}
\end{figure}

Figure~\ref{fig:prompt_for_evaluation} depicts the prompt used for Llama 2-Chat-70B to assess speech-style strength. In this prompt, we specify the text style transfer task and emphasize and elucidate the features that warrant special attention, aiding the model in accurate evaluations.

\section{Human Annotation}
To improve consistency among evaluators, we use reference examples to reduce the level of abstraction for complex sub-styles, shown in Figures~\ref{fig:human_annotation_correlation}, \ref{fig:human_annotation_correlation-3-2}.

\label{sec:human_annotation}
\subsection{Human Annotation for Prior Fine-grained Analysis}
\label{sec:human_annotation_for_Prior_Fine-grained_Analysis}
\begin{figure}
    \centering
    \includegraphics[scale=0.45]{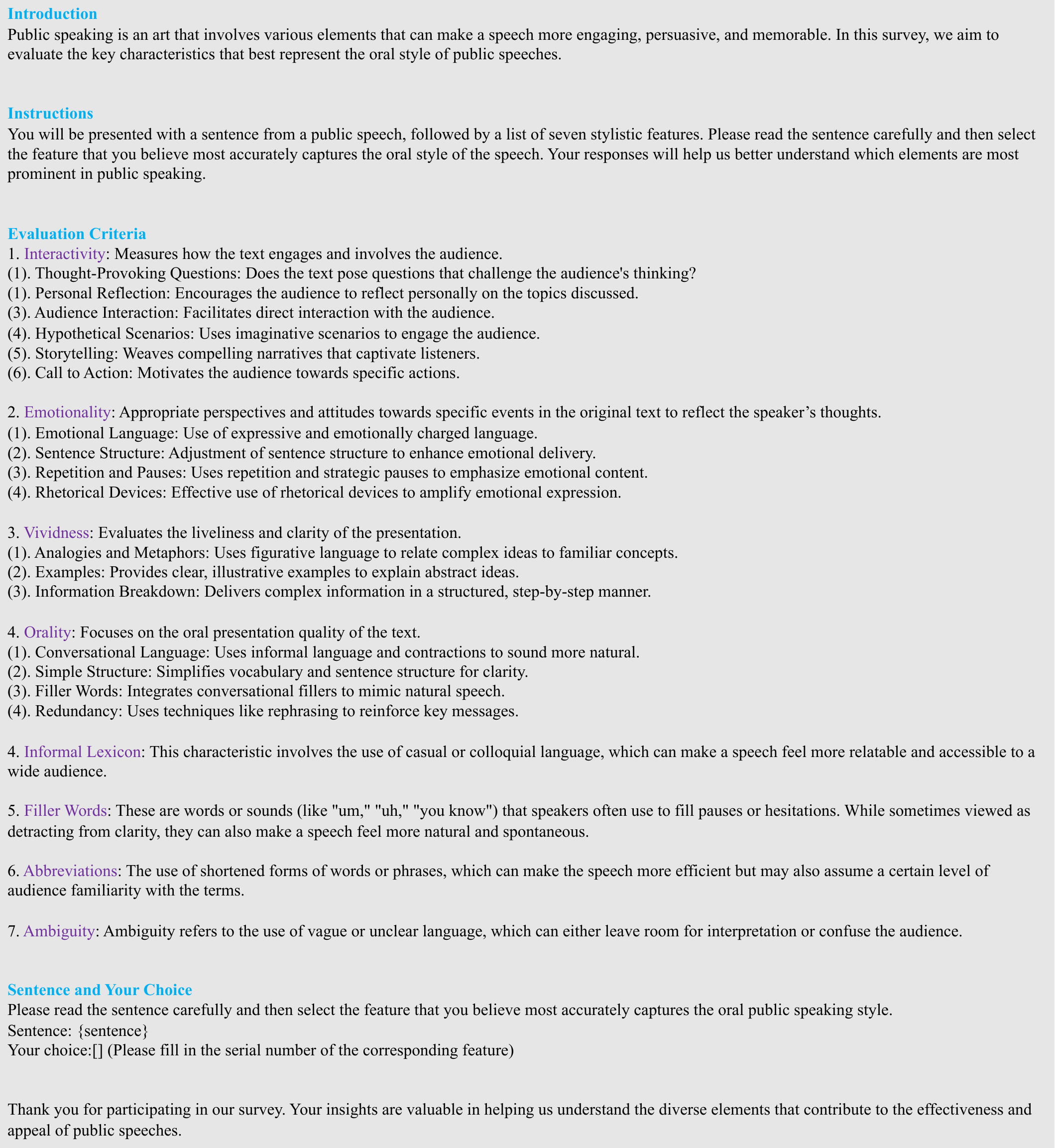}
    \caption{Human questionnaire for Prior Fine-grained Analysis.}
    \label{fig:human_annotation_correlation-3-2}
\end{figure}

For the annotation of real-world public speaking data features, we sample 300 sentence instances from four public speaking contexts: TED Talks, political speeches, academic presentations, and lecture transcripts. Based on prior research on oral public speaking~\citep{mccroskey2003principles,Beebe2005ThePS,Atkinson_1985,Halliday1989SpokenAW}, we provide seven candidate features for annotation: (1) Interactivity, (2) Emotionality, (3) Filler Words, (4) Vividness, (5) Ambiguity, (6) Abbreviations, and (7) Informal Lexicon. Each annotator is required to perform two types of annotations: (1) Multi-label: annotators are required to select all candidate features contained in each instance. (2) Best-one label: annotators are required to choose the feature that most strongly represents the public speaking style of the sentence. Definitions for each candidate dimension provided to the annotators are as follows, and the full questionnaire is shown in Figure~\ref{fig:human_annotation_correlation-3-2}, where we further provide some detailed guidelines or examples to reduce the abstraction of sub-styles for a more accurate annotation:
\begin{itemize}
\item{\textbf{Interactivity:} Interactivity in public speaking refers to the speaker engaging with the audience through various means such as posing thought-provoking, facilitating personal reflection, and crafting intriguing hypothetical scenarios.}
\item{\textbf{Emotionality:} Public speaking contains the speaker's appropriate views and attitudes on specific events to reflect the speaker's emotional tendencies and inner thoughts.}
\item{\textbf{Filler Words:} Filler words include "um," "ah," "you know," and similar phrases that speakers use to fill pauses during their speech. While sometimes seen as a sign of nervousness, they can also function to give the speaker time to think.}
\item{\textbf{Vividness:} In public speaking, speakers should present information in a lively, easy-to-understand way, such as using analogies and metaphors to make complex ideas more accessible and engaging.}
\item{\textbf{Ambiguity:} This feature reflects the overall clarity of the sentence. Ambiguity can arise from using complex, unclear, or overly verbose language that makes the content difficult to follow. 
}
\item{\textbf{Abbreviations:} This includes the use of shortened forms of words or phrases. In speeches, abbreviations can make communication quicker and fit more informal settings.}
\item{\textbf{Informal Lexicon:} Public speakers tend to use casual or colloquial language. It contrasts with formal speech and can make the speaker appear more relatable and approachable.}
\end{itemize}

\subsection{Human Annotation for Text-Level Style Strength Evaluation}
\label{sec:annotation_for_Text-Level_Style_Strength_Evaluation}
\begin{figure}
    \centering
    \includegraphics[scale=0.45]{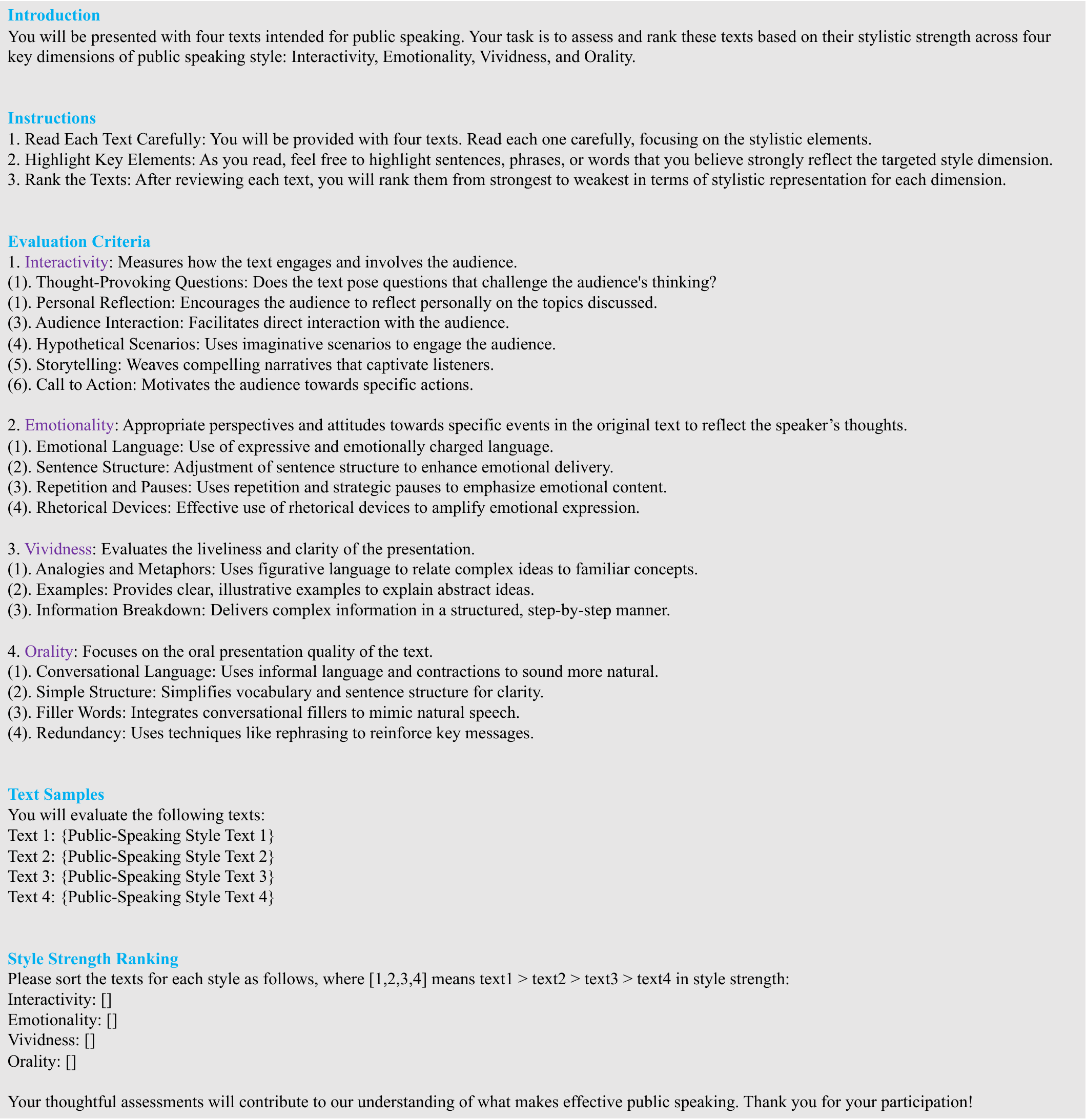}
    \caption{Human questionnaire for public-speaking style strength annotation.}
    \label{fig:human_annotation_correlation}
\end{figure}

Figure~\ref{fig:human_annotation_correlation} presents the questionnaire designed for
human annotators tasked with public-speaking style strength annotation mentioned in Section~\ref{sec:correlation}. We engaged 3 graduate students who are proficient in English but are not linguists as annotators, achieving inter-annotator consistency of 0.81631 (Krippendorff's $\alpha$ coefficient~\citep{Krippendorff2011ComputingKA}).

\subsection{Human Annotation for Semantic Preservation}
\label{sec:annotation_for_Semantic_Preservation}
We conducted the following two experiments, demonstrating that the QA-based method shows a strong correlation with human evaluations in semantic preservation evaluation and has the advantage of detecting and locating missing information in long text style transfer tasks.
\subsubsection{Experiment 1: Semantic preservation methods comparison}
\paragraph{Experimental Setup} We sample 20 sets of texts, each consisting of one original text and three stylized texts generated by different models, with each text having a token length between 300 and 500 (To accommodate the 512-token context limit of BLUERT and BertScore). We ask three evaluators to rank the three stylized texts based on semantic consistency, achieving inter-annotator consistency of 0.7693 ( Krippendorff's $\alpha$).
\paragraph{Automatic Evaluation Methods}
\begin{enumerate}
    \item \textbf{BertScore}~\citep{zhang2019bertscore}: BERTScore leverages the pre-trained contextual embeddings from BERT and matches words in candidate and reference sentences by cosine similarity.
    \item \textbf{BLUERT}~\citep{bleurt}: BLUERT is built using multiple phases of transfer learning starting from a pretrained BERT model. It has a good correlation with human judgments of semantic consistency.
    \item \textbf{QA-based(ours)}: The method proposed in Section~\ref{sec:semantic_preservation_eval} to evaluate the semantic consistency of long documents from two parts: key information and logical structure.
\end{enumerate}

\paragraph{Results}
As shown in Table~\ref{tab:Results_of_Correlation_Semantic_preservation}, the experimental results indicate that the QA-based method proposed in this paper shows a strong correlation with human evaluations in assessing semantic consistency, outperforming model-based evaluation methods such as BELURT and BertScore.

\subsubsection{Experiment 2: Human evaluation on QA pairs}

We sample 100 incorrect QA pairs from Section~\ref{sec:exps} "Public Speaking Ability Evaluation", which are correct when based on the original text, and incorrect when based on the stylized text. Then, we manually check whether the key information in these QA pairs remained consistent between the original and stylized texts. Results show that \textbf{87/100} QA pairs successfully detected the corresponding knowledge missing in the text after style transfer.

\subsubsection{Experiment 3: QA on Questions generated based on Public-Speaking Style Texts}
\paragraph{Experimental Setup} we sampled 20 Public-Speaking Style texts generated by GPT-3.5 and created 20 questions for each text by GPT-4, totaling 400 QA pairs (using the method mentioned in Section~\ref{sec:semantic_preservation_eval}). The QA model is Llama-3-8B-Instruct and the input includes Public-Speaking Style texts and the corresponding QA questions.
\paragraph{Results} The accuracy of the QA model’s responses was 98\%, which indicates that the model has a robust QA capability for handling Public-Speaking Style texts. The above results further indicate that the low accuracy of the QA model based on stylized texts is caused by semantic loss rather than the inability of the model to process Public-Speaking Style texts.

\section{Case Study}
\subsection{Audio and video demos of PSST}
\label{sec:demo}
\begin{figure}
    \centering
    \includegraphics[scale=0.45]{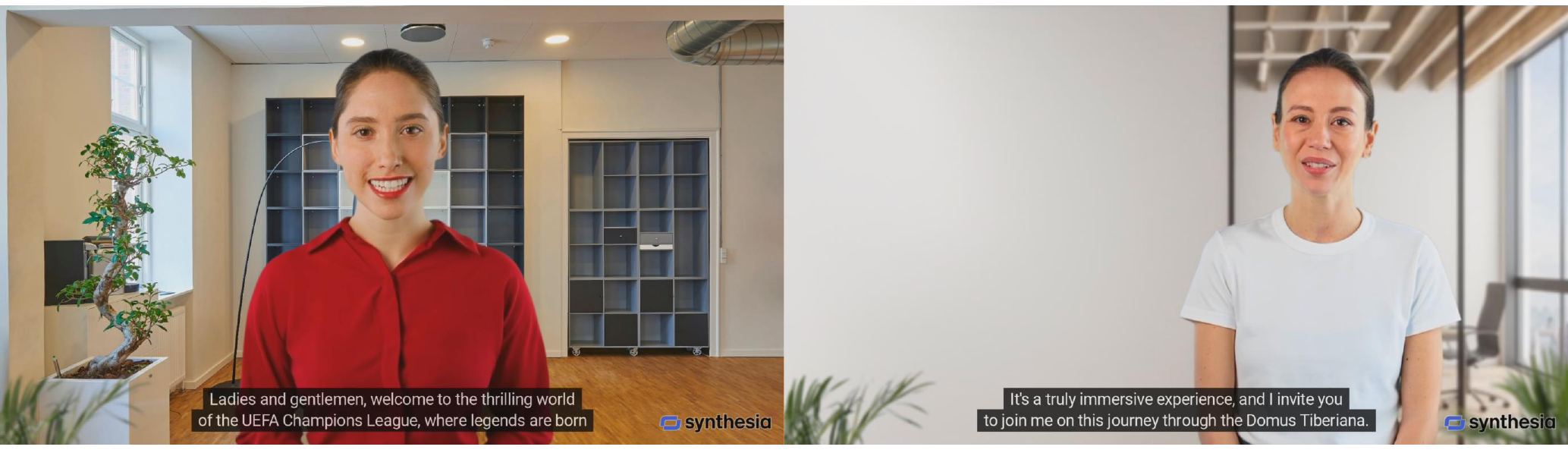}
    \caption{Screenshots of videos based on LLM-generated public speaking style text. All demos are available at this \href{https://grateful-sesame-4aa.notion.site/Presentation-of-PSST-de0bcc31121442278a158851aa180fdf}{URL}.}
    \label{fig:video_demo}
\end{figure}

To vividly demonstrate the PSST task, we utilize text-to-speech\footnote{https://elevenlabs.io/} and text-to-video\footnote{https://www.synthesia.io/} tools to animate virtual avatars delivering public speaking based on text generated by LLMs. By comparing speech and video based on text before and after style transfer, we provide a more intuitive illustration of the practical value and applicability of the PSST task. Figures~\ref{fig:video_demo} are screenshots of the generated video. All demos are available at this URL\footnote{https://grateful-sesame-4aa.notion.site/Presentation-of-PSST-de0bcc31121442278a158851aa180fdf}.


\subsection{Bad Cases of LLMs in PSST}
\label{Bad_case_of_LLMs_in_PSST}
\begin{figure}
    \centering
    \includegraphics[scale=0.45]{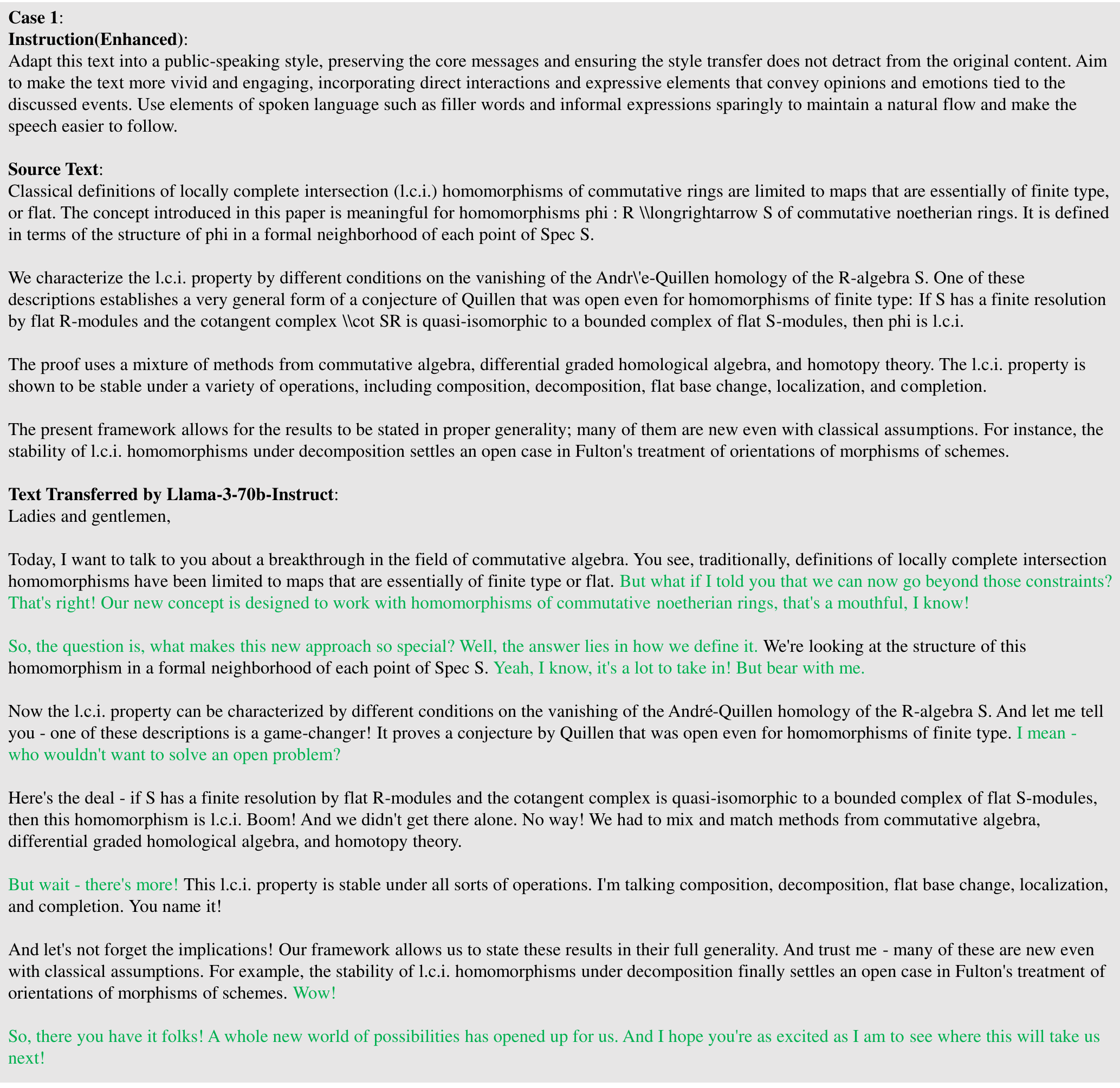}
    \caption{Case of Llama-3-70b-Instruct in PSST with enhanced instruction. The \textcolor{green}{green} parts indicate strong stylistic strength. The transferred text is more vivid, emotional, and attractive to the audience than the source text.}
    \label{case_1}
\end{figure}

\begin{figure}
    \centering
    \includegraphics[scale=0.45]{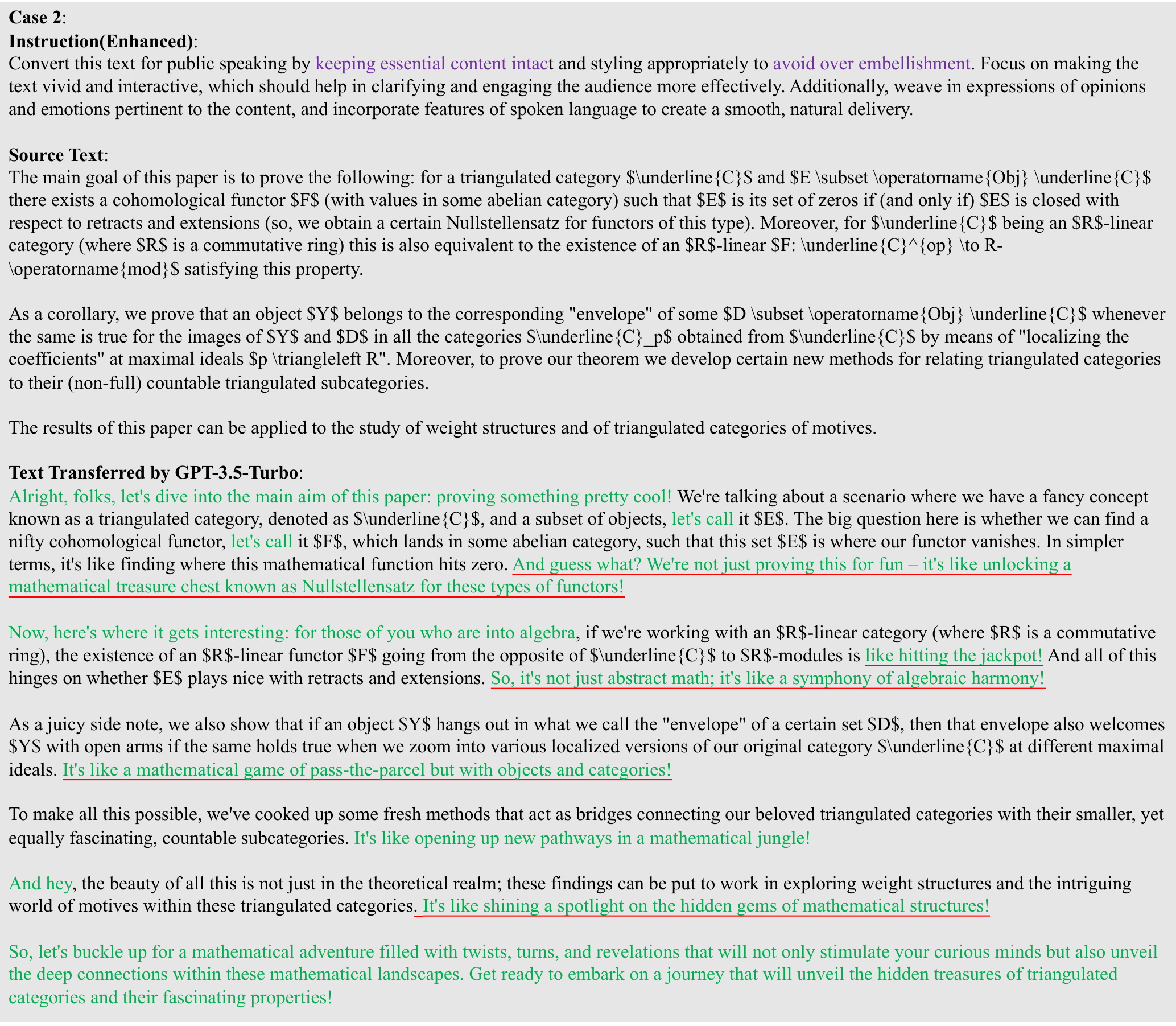}
    \caption{Case of GPT-3.5 in PSST with enhanced instruction. The \textcolor{green}{green} parts indicate strong stylistic strength. The transferred text uses many metaphors and reflects rich emotions. However, the stylization of the \textcolor{red}{underlined} section seems to be intentionally exaggerated and appears unnatural, diverging from human preferences.}
    \label{case_2}
\end{figure}

\begin{figure}
    \centering
    \includegraphics[scale=0.45]{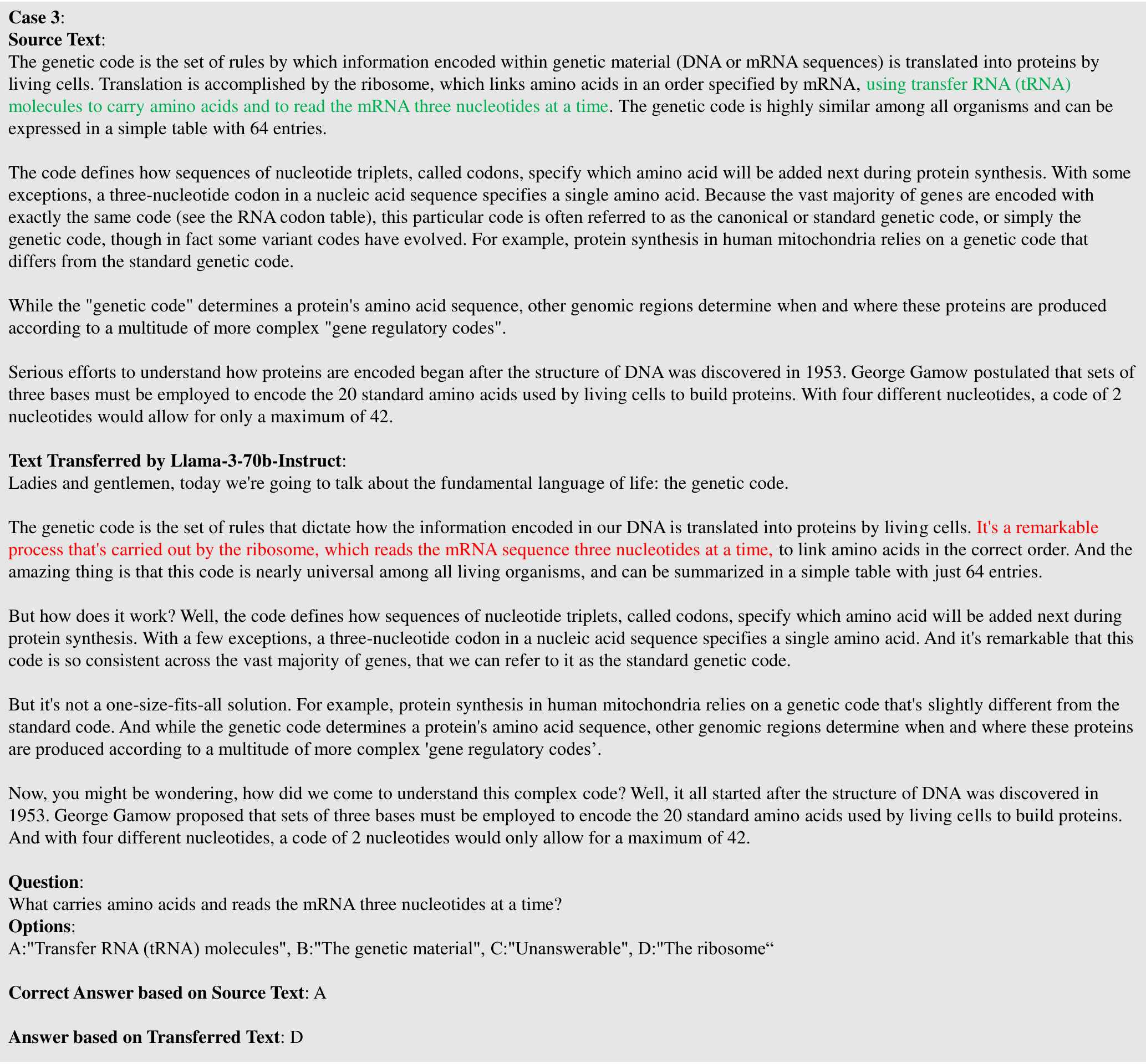}
    \caption{Case of evaluating the semantic preservation of Llama-3-70b-Instruct in PSST. The \textcolor{green}{green} part indicates the location of the answer to the question in the source text. The \textcolor{red}{red} part indicates the location corresponding to the text after the transfer, but the key information of the question is lost.}
    \label{case_3}
\end{figure}

\begin{figure}
    \centering
    \includegraphics[scale=0.45]{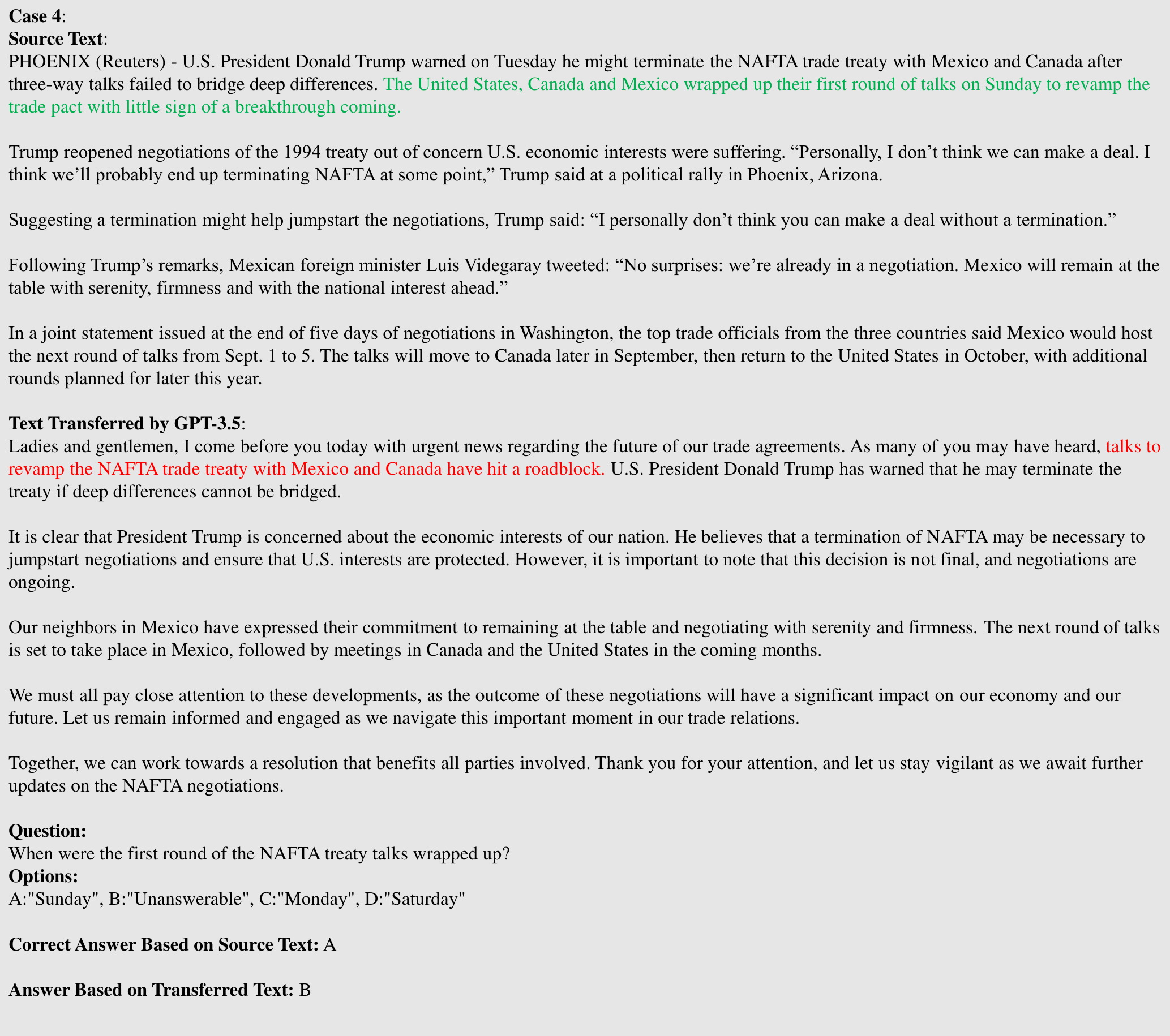}
    \caption{Case of evaluating the semantic preservation of GPT-3.5 in PSST. The \textcolor{green}{green} part indicates the location of the answer to the question in the source text. The \textcolor{red}{red} part indicates the location corresponding to the text after the transfer, but the key information of the question is lost.}
    \label{case_4}
\end{figure}

In Figures~\ref{case_1},~\ref{case_2},~\ref{case_3},~\ref{case_4}, we provide specific cases of challenges encountered by LLMs during the execution of PSST, as discussed in Section~\ref{sec: evaluation_results}. These cases encompass issues denoted as Excessive Style Strength, Uneven Style Strength Distribution, and Severe Semantic Degradation.

\begin{figure}
    \centering
    \includegraphics[scale=0.45]{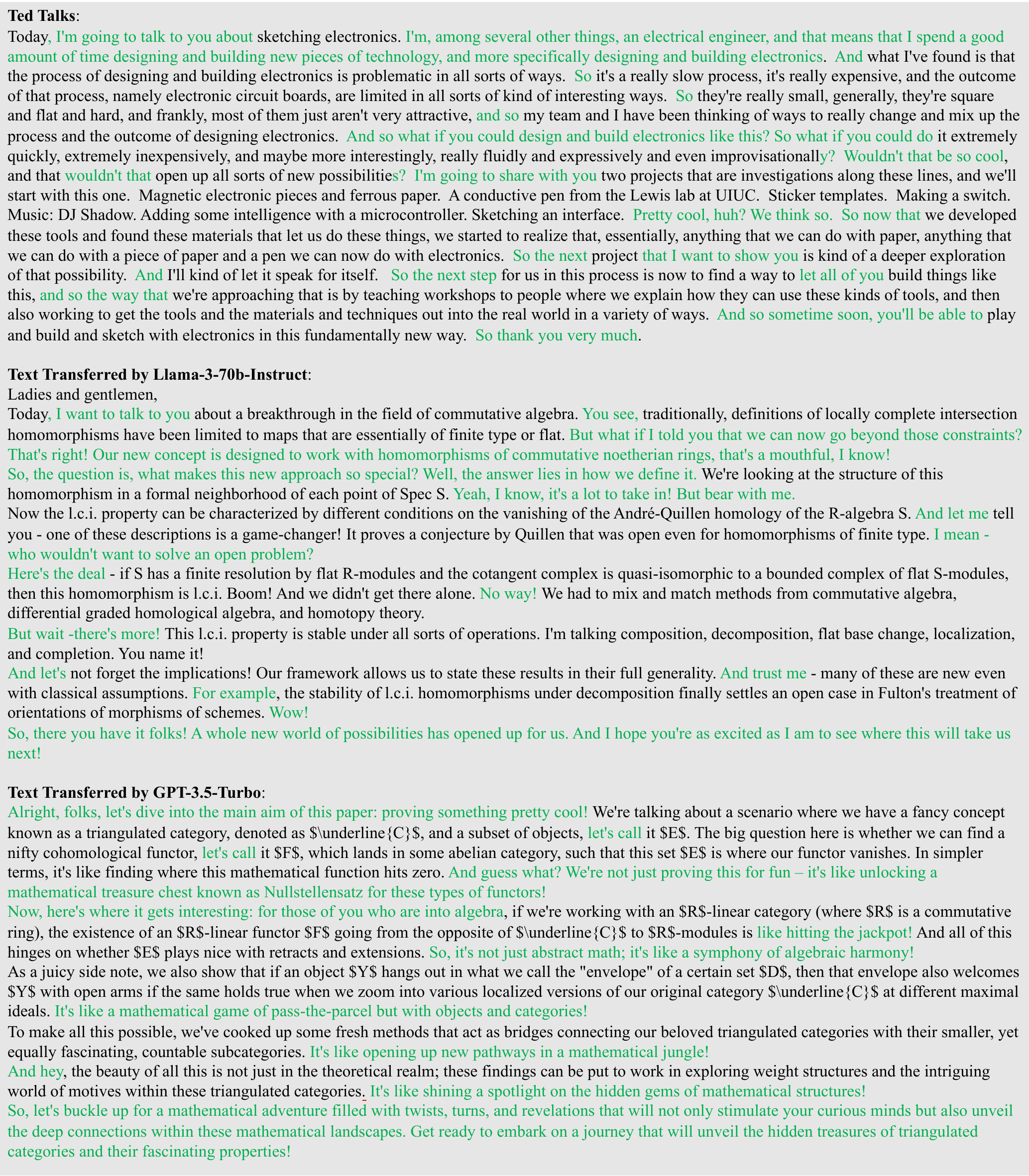}
    \caption{Comparison of different public-speaking style texts (Ted Talks, Llama-3-70b-Instruct, GPT-3.5-Turbo). The \textcolor{green}{green} parts indicate strong stylistic strength. In TED talks, stylistic elements are uniformly distributed, whereas in texts generated by LLMs, these elements tend to concentrate disproportionately at the beginning and the end. Moreover, LLAMA-3 generates moderately stylized texts, while GPT-3.5 tends to produce texts that are excessively stylized.}
    \label{case_0}
\end{figure}

\subsection{Bad Cases of Llama 2-Chat-70B in Speech-Style Strength Evaluation}
\label{Bad_case_of_Llama2_in_Evaluation}
\begin{figure}
    \centering
    \includegraphics[scale=0.45]{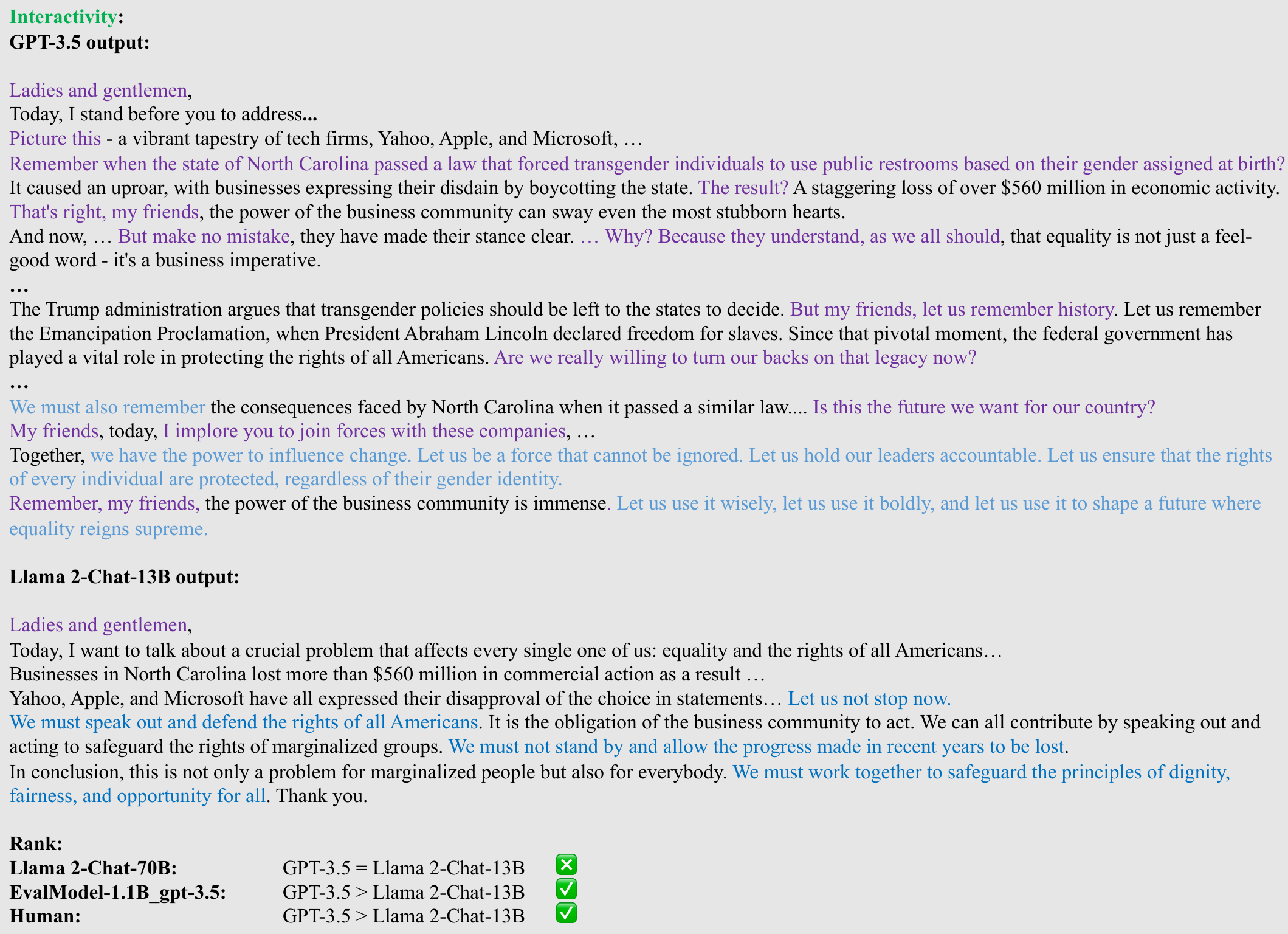}
    \caption{An instance of the evaluation failure in the interactivity dimension using  \textbf{Llama 2-Chat-70B}. The provided prompt is shown in Figure~\ref{fig:prompt_for_evaluation}. The color code represents human preferences, with \textcolor{purple}{purple} indicating a stronger preference and \textcolor{blue}{blue} indicating a weak preference. Notably, LLaMA2-Chat-70B assigns the same ranking to both texts, highlighting a failure in speech-style strength evaluation.}
    \label{evaluation_case}
\end{figure}

Inconsistency and insensitivity to the style strength of LLMs as speech-style strength refers to the challenge of obtaining inconsistent results for the same input across multiple tests and assigning the same score despite noticeable variations in speech-style strength. Examples are shown in Figure~\ref{evaluation_case}.

\subsection{Bad Cases of ChatGPT in Content Preservation Evaluation}
\label{Bad_Cases_of_ChatGPT_in_Content_Preservation_Evaluation}

\begin{figure}
    \centering
    \includegraphics[scale=0.45]{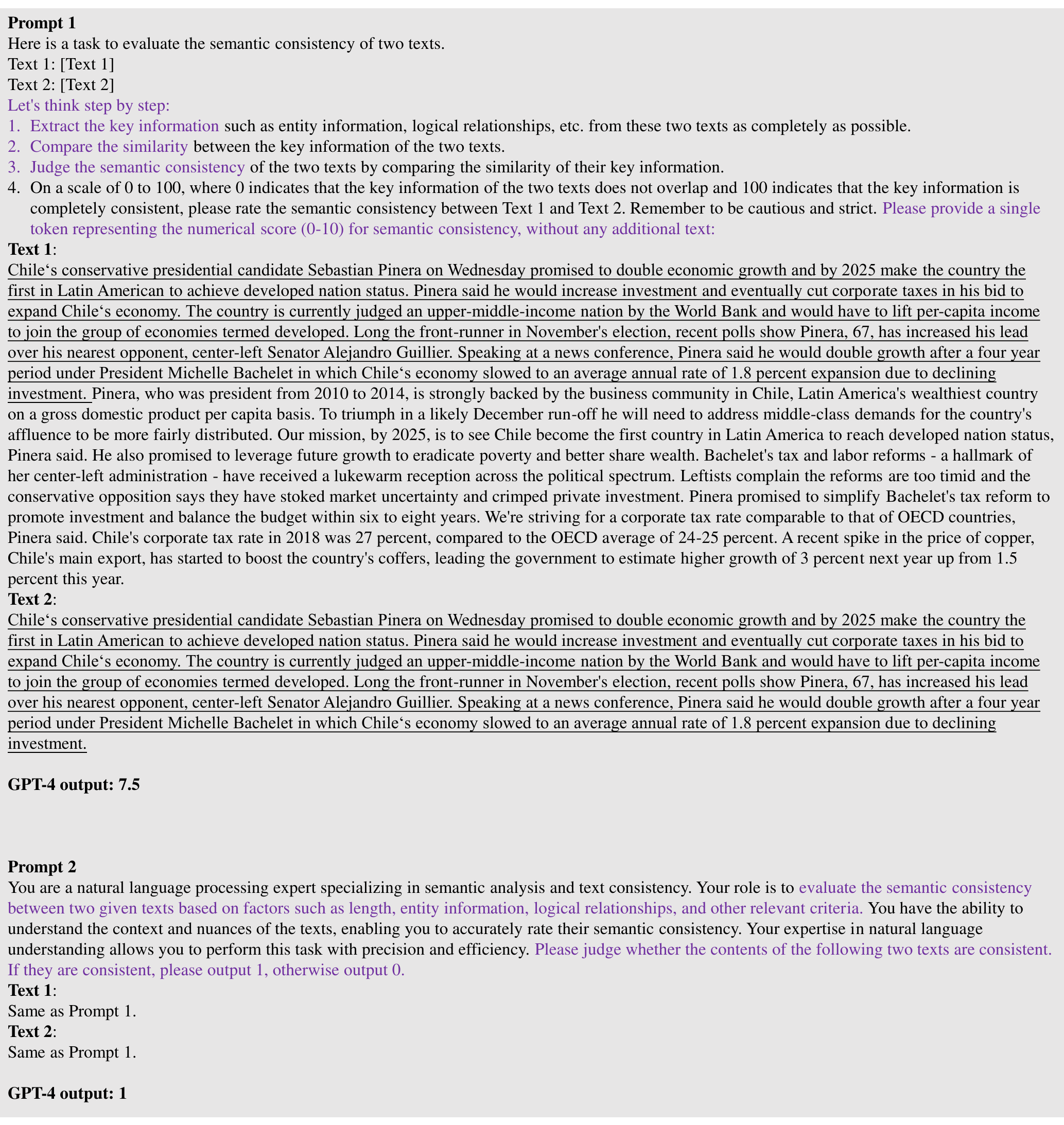}
    \caption{Prompts and results of GPT-4 evaluating the semantic consistency of two texts. Text 2 is part of Text 1, and the length of Text 2 is about 60\% less than Text 1. GPT-4 scored the semantic consistency of the two texts as 7.5 (on a scale of 0-10) and 1 (0 for inconsistency and 1 for consistency), which indicates that GPT-4 cannot evaluate the semantic consistency between two long texts.}
    \label{yyyzx}
\end{figure}

We attempt to directly evaluate the semantic consistency of two passage-level texts using GPT-3.5 and GPT-4. We designed a variety of prompts including those shown in Figure~\ref{yyyzx}. When two input texts are completely unrelated, GPT can correctly determine that they are not related. But when a text is part of another text, GPT will tell that their semantics are consistent. Whether we ask GPT to score semantic consistency or let it judge whether two texts are consistent, their performance is unsatisfactory.

\end{document}